%% file: main_CSDN.tex
\documentclass[10pt,journal,cspaper,compsoc]{IEEEtran}
\ifCLASSINFOpdf
\else
\fi
\usepackage{amssymb}

\usepackage{epsfig}
\usepackage{epstopdf}
\usepackage{subfigure}
\usepackage{amsmath}
\usepackage{booktabs}
\usepackage{tabularx}
\usepackage{times}
\usepackage{graphicx}
\usepackage{amssymb}

\usepackage{color}
\usepackage{cite}
\usepackage{psfig}
\usepackage{algorithm}
\usepackage{algorithmic}
\usepackage{graphicx}
\usepackage{cases}
\usepackage{multirow}

\usepackage{psfig}
\usepackage{algorithm}
\usepackage{algorithmic}
\usepackage{graphicx}
\usepackage{amsmath}
\usepackage{booktabs}
\usepackage{subfigure}
\usepackage{cases}
\usepackage{enumerate}
\usepackage{wrapfig}
\usepackage{picinpar}
\usepackage{pbox}

\newcommand{\thickhline}{%
	\noalign {\ifnum 0=`}\fi \hrule height 1pt
	\futurelet \reserved@a \@xhline
}

\begin{document}
%
\title{CSDN: Cross-modal Shape-transfer Dual-refinement Network for Point Cloud Completion}

\author{
\IEEEauthorblockN{Zhe Zhu, Liangliang Nan, Haoran Xie, \textit{Senior Member, IEEE}, Honghua Chen, Jun Wang, Mingqiang Wei, \textit{Senior Member, IEEE}, and Jing Qin} \\
\thanks{Z. Zhu, H. Chen, J. Wang and M. Wei are with the School of Computer Science and Technology, Nanjing University of Aeronautics and Astronautics, Nanjing, China (e-mail: zhuzhe0619@nuaa.edu.cn;  chenhonghuacn@gmail.com; wjun@nuaa.edu.cn; mingqiang.wei@gmail.com).}
\thanks{L. Nan is with the Urban Data Science Section, Delft University of Technology, Delft, Netherlands (e-mail: liangliang.nan@tudelft.nl).}
\thanks{H. Xie is with the Department of Computing and Decision Sciences, Lingnan University, Hong Kong, China (e-mail: hrxie2@gmail.com).}
\thanks{J. Qin is with the School of Nursing, The Hong Kong Polytechnic University, Hong Kong, China (e-mail: harry.qin@polyu.edu.hk). }
}

%



\IEEEtitleabstractindextext{%
\begin{abstract}
How will you repair a physical object with some missings? 
You may imagine its original shape from previously captured images, recover its overall (global) but coarse shape first, and then refine its local details. 
We are motivated to imitate the physical repair procedure to address point cloud completion.
To this end, we propose a cross-modal shape-transfer dual-refinement network (termed CSDN), a coarse-to-fine paradigm with images of full-cycle participation,  for quality point cloud completion.
CSDN mainly consists of ``shape fusion" and ``dual-refinement" modules to tackle the cross-modal challenge.
The first module transfers the intrinsic shape characteristics from single images to guide the geometry generation of the missing regions of point clouds, in which we propose IPAdaIN to embed the global features of both the image and the partial point cloud into completion. The second module refines the coarse output by adjusting the positions of the generated points, where the local refinement unit exploits the geometric relation between the novel and the input points by graph convolution, and the global constraint unit utilizes the input image to fine-tune the generated offset.
Different from most existing approaches, CSDN not only explores the complementary information from images but also effectively exploits cross-modal data in the \textit{whole} coarse-to-fine completion procedure.
Experimental results indicate that CSDN performs favorably against twelve competitors on the cross-modal benchmark.

\end{abstract}

}

\maketitle

\IEEEdisplaynontitleabstractindextext

%
\IEEEpeerreviewmaketitle

%
%
%
%

\input{intro}
\label{secIntroduction}

\input{related_work}

\label{secRelatedWork}
\input{method}

\label{secOverview}

\input{experiment}

\label{secEXP}

\input{Feature_Visualization}

\label{secFeaVis}

\input{limitation}
\label{secLim}

\input{conclusion}

\label{secCon}


\ifCLASSOPTIONcaptionsoff
  \newpage
\fi

\bibliographystyle{IEEEtran}
\bibliography{pcf}

\end{document}

%% file: intro.tex
\section{Introduction}
\begin{figure}[ht]
  \includegraphics[width=\linewidth]{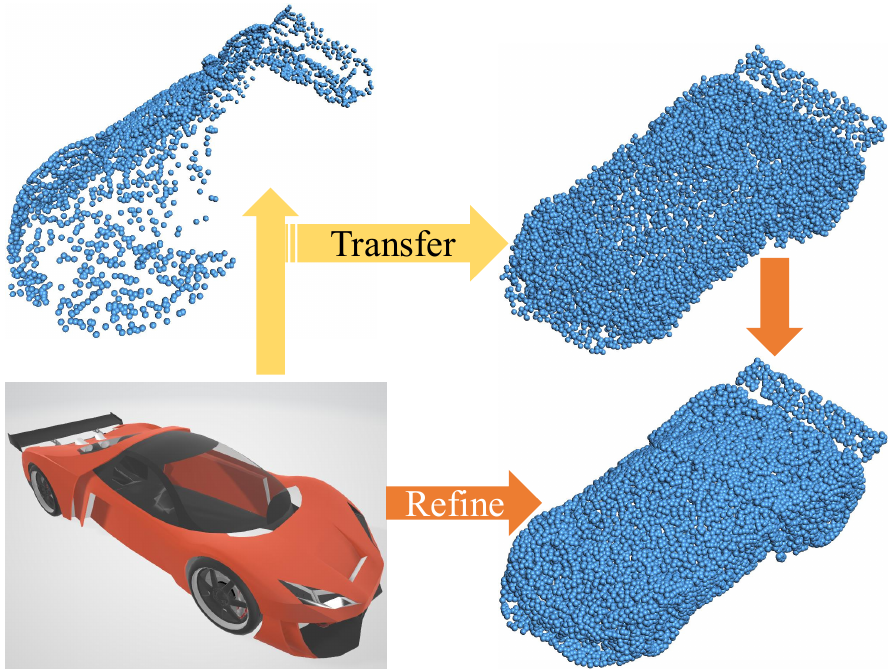}
  \caption{Image is an important modality to provide complementary information (both structure and detail information) for 3D (sparse) data. Inspired by how a man/woman repairs physical objects, we consider that the image should participate in the whole procedure of coarse-to-fine repair. Differently from existing efforts in this topic, CSDN is a new point cloud completion paradigm that exploits cross modalities in the \textit{whole} coarse-to-fine completion procedure (Yellow arrows: the shape fusion stage; Orange arrows: the dual-refinement stage).}
  \label{fig:teaser}
\end{figure}

As a commonly used representation, point clouds express the geometry of objects in 3D space with a simple and flexible data structure \cite{corr/abs-2205-07417}. 
The raw point clouds, which are often captured by scanning devices (e.g., laser scanners), unavoidably have missing regions. Some of main reasons include the inter-object occlusion, self-occlusion, surface reflectivity, and limited scanning range~\cite{corr/abs-2203-03311}. 
It is possible to complete a partial point cloud by carefully positioning 3D scanners to carry out multi-pass scanning. However, such remedial measures are time-consuming and often hopeless due to the limited scene accessibility. Also, it is promising to learn a completion mapping with single-modal point clouds. Two challenges of using the single-modal point clouds will be encountered. First, the low-scanning resolution of 3D sensors (e.g., Kinect) often leads to the sparseness of captured data. Determining whether an arbitrary missing region originates from incompleteness or intrinsic spareness is very difficult in such sparse point clouds. Second, the inference uncertainty of missing regions will happen due to the limited geometry cues available \cite{zhang2021view}.


Cross-sensors (e.g., depth cameras and camera-LiDAR scanners) are now accessible and affordable to continuously and simultaneously capture color images and point clouds of 3D scenes. Scene images have two main merits: (1) compared to 3D data acquisition using static/mobile laser scanners, images of the scene or in particular the concerned missing regions are easier to acquire \cite{cgf/NanSC14}; (2) images have high resolutions and rich textures, potentially facilitating various geometric processing tasks by providing complementary information to point clouds \cite{tomccap/WangXWLW22}. Although recent years have witnessed considerable efforts of using images to help understand point clouds, it is still challenging to fuse data from the cross sensors, making them `really' complementary for quality geometric processing tasks. 


To address the cross-modal challenge in point cloud completion, one can absorb the wisdom of physical object repair. Imagine how a professional restorer repairs a physical object with some missings.
She/he will perceive the object's original shape from previously captured images and first recover its global yet the coarse shape and then refine its local details. We attempt to imitate the physical repair procedure to design a novel cross-modal point cloud completion paradigm. 
To this end, we propose a cross-modal shape-transfer dual-refinement network (termed CSDN), which exploits the complementary information of images and input partial geometry for quality point cloud completion.

CSDN is designed to fuse the features of both the image and point cloud based on their role in conveying a complete shape. For effective learning of both global and local features at different completion stages, we exploit different strategies to tackle the cross-modal challenge. To fully integrate image information for shape completion in a coarse-to-fine manner, CSDN consists of a ``shape fusion" module to generate a coarse yet complete shape and a    ``dual-refinement" module to obtain the completion result.

First, 
rather than generating the complete model with only single modality data, we develop a novel solution called IPAdaIN (a variant of Adaptive Instance Normalization \cite{huang2017arbitrary,karras2019style}) to embed the global features of both the images and the partial point cloud. To fuse the information from the two modalities, IPAdaIN adaptively transfers the ``shape style" (i.e., the global shape information) obtained from the image to the point cloud domain by feature normalization, which facilitates obtaining an initial complete 3D shape with the aid of images.

Then, the dual-refinement module takes the generated point cloud as input and further refines the shape details. Specifically, we design this module with a pair of refinement units, called local refinement and global constraint, respectively. To recover geometric details, we apply graph convolution in the \emph{local refinement} unit. The \emph{global constraint} unit leverages image features to constrain the offset generated by \emph{local refinement}. This dual-path refinement strategy enables CSDN to precisely capture both global shape structures and local details. 

CSDN explores the complementary structure information from images, and it synergizes both point and image features in the whole coarse-to-fine completion procedure, as shown in Figure~\ref{fig:teaser}. This makes CSDN distinct from the recently proposed ViPC~\cite{zhang2021view}.
ViPC also exploits an additional single-view image to provide the global structural priors in its coarse-completion stage. However, it cannot recover high-frequency details and local topology for complex shapes in the refinement stage since the features learned from different modalities are simply concatenated. As a result, the completed shapes from ViPC are often noisy and lack finer geometric details. 
In contrast, our CSDN enables the recovery of finer details and topology by disentangling features for \emph{local refinement} and features for \emph{global constraint}.

Extensive experiments clearly show the superiority of CSDN over its ten competitors. For example, CSDN reduces the Mean Chamfer Distance by 0.281 compared to PoinTr \cite{yu2021pointr}, and it outperforms ViPC \cite{zhang2021view} by a large margin of 22.3\% on the benchmark dataset ShapeNet-ViPC \cite{zhang2021view}.
The main contributions can be summarized as follows.
\begin{itemize}
  \item By imitating the physical repair procedure, we propose a novel cross-modal shape-transfer dual-refinement network for point cloud completion. Compared with previous cross-modal methods, CSDN applies the disentangled feature fusion strategy, which significantly improves the completion performance.
  \item We propose a shape fusion module for generating a coarse yet complete shape, in which a novel solution called IPAdaIN is designed to adaptively transfer the global shape information obtained from the image to the point cloud domain by feature normalization. 
  \item We propose a dual-refinement module, enabling CSDN to capture both global shape structures and local details.
\end{itemize}

\begin{figure*}[ht]
  \centering
  \includegraphics[width=\textwidth]{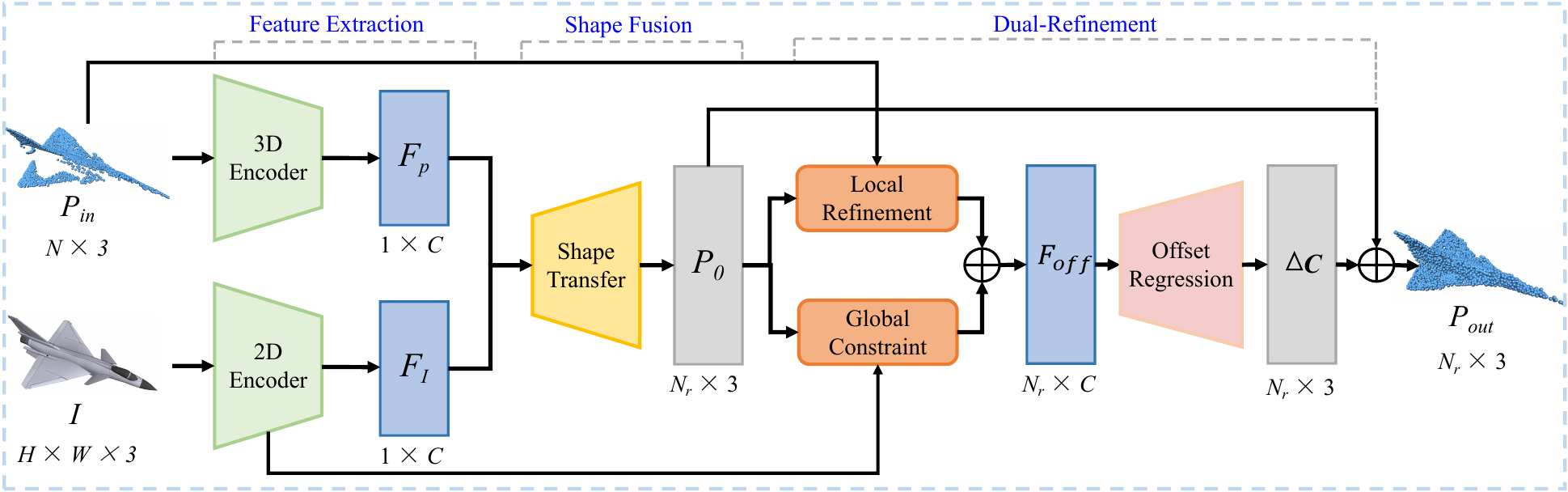}
\caption{The architecture of Cross-modal Shape-transfer Dual-refinement Network (CSDN). CSDN is designed as a coarse-to-fine structure, and complementary images participate in the whole process. CSDN consists of three modules: feature extraction, shape fusion, and dual refinement. Different from the existing wisdom, cross-modal data are involved in the whole coarse-to-fine completion procedure of CSDN, ensuring the completion result with both rich structures and details.}
  \label{fig:overview}
\end{figure*}

%% file: related_work.tex
\section{Related Work}
This section reviews recent advances in learning-based point cloud analysis, point cloud generation, single-view reconstruction, and 3D shape completion, followed by cross-modal feature fusion.

\subsection{Point Cloud Analysis}
Most of recent neural networks for point cloud analysis are designed to process unordered point clouds directly, rather than converting them into 2D grids or 3D voxels in advance. As the pioneer, PointNet~\cite{qi2017pointnet} proposes to directly process points through MLPs and uses a symmetric function to aggregate features. Then, PointNet++~\cite{qi2017pointnet++} utilizes a hierarchical architecture to capture local information. To generalize typical CNN to point clouds, PointCNN~\cite{li2018pointcnn} learns a transformation matrix for local points permutation. PointConv~\cite{wu2019pointconv} and KPConv~\cite{thomas2019kpconv} build convolution weights with point coordinates. By expressing a point cloud using a graph structure, DGCNN~\cite{wang2019dynamic} proposes EdgeConv for efficient local geometry learning.
Several follow-up works~\cite{lin2020convolution,pistilli2020learning,iccv/ZhouFFW0L21,wei2022agconv} further demonstrate the efficiency of graph representations for local point clouds. More recently, Point Transformer~\cite{zhao2021point} and PCT~\cite{guo2021pct} extend the Transformer architecture to analyze point clouds. These works greatly facilitate the development of point cloud processing in downstream tasks.

\subsection{Point Cloud Generation}
Point cloud generation aims to learn generative models on 3D shapes. rGAN~\cite{achlioptas2018learning} first proposes to use GAN to generate point clouds, followed by some variants~\cite{shu20193d,li2021sp,tang2022warpinggan} that aim at generating higher quality point clouds by exploiting local structure information. As a representative flow-based method, PointFlow~\cite{yang2019pointflow} treats shapes and shape points as a two-level hierarchy of distributions. To better characterize the data distribution, ShapeGF~\cite{cai2020learning} generates points by learning the stochastic gradient of its log density field. Generative methods are also widely adopted in other 3D vision tasks. For example, GSPN~\cite{yi2019gspn} generates proposals with an analysis-by-synthesis strategy. More recently, PointGrow~\cite{sun2020pointgrow} and AutoSDF~\cite{mittal2022autosdf} use auto-regressive models to generate 3D shapes. PointGrow learns point distribution in a point-by-point manner. AutoSDF employs Truncated-Signed Distance Field (T-SDF) as the 3D representation and utilizes the transformer architecture to learn non-sequential prior over 3D shapes. Note that the goal of most of the above methods is to generate high-fidelity 3D shapes by learning and mimicking existing data distributions. Different from them, Point cloud completion methods~\cite{yuan2018pcn,yu2021pointr,zhang2021view} and point cloud upsampling methods~\cite{yu2018pu,qian2021pu,li2021point} are conditioned by the input partial/sparse point clouds. These efforts focus on mapping and completing input partial shapes to their complete versions.

\subsection{Shape Completion}
Traditional methods are often geometry- or template-based. The geometry-based methods either fill small holes in 3D models by local interpolations \cite{berger2014state,davis2002filling,nealen2006laplacian}, or infer the complete shapes by exploiting structure regularities, such as symmetry and repetitive patterns \cite{mitra2006partial,mitra2013symmetry,sung2015data,thrun2005shape}. The template-based methods \cite{blanz1999morphable,chauve2010robust,han2008bottom,kalogerakis2012probabilistic} complete the geometry by matching the partial shape with template models from a database. However, for this conventional wisdom
of shape completion, users have to tweak parameters multiple
times to obtain satisfied results in practical scenarios. This inconvenience heavily discounts the efficiency and user experience.

Early learning-based methods \cite{dai2017shape,han2017high,xie2020grnet} utilize voxel-based representations for 3D convolutional neural completion networks, but suffering from expensive computation cost.
Later, the PointNet-based encoder-decoder architecture \cite{qi2017pointnet} inspires many new works. For example, the pioneering one, PCN \cite{yuan2018pcn}, generates points in a coarse-to-fine manner and folds the 2D grids to reconstruct a complex complete shape. This coarse-to-fine completion framework has also been employed by~\cite{xie2020grnet,zhang2020detail}.
TopNet \cite{tchapmi2019topnet} introduces a hierarchical rooted tree structure decoder to generate points of different levels of details.
The following work MSN \cite{liu2020morphing} tackles the completion problem by generating different parts of shape with a residual sub-network to refine the coarse points. 
PF-Net \cite{huang2020pf} extracts multi-scale features to encode both local and global information and then fuses them to reconstruct complete point clouds. Besides, the adversarial training strategy is used to further enhance the point quality \cite{huang2020pf}. 
Also, Wang et al.~\cite{wang2020cascaded} propose to synthesize the dense and complete object shapes in a cascaded refinement manner, and jointly optimize the reconstruction loss and an adversarial loss.
RL-GAN-Net~\cite{sarmad2019rl} first tries to use an RL agent to drive a generative adversarial network to predict a complete point cloud.
PMP-Net \cite{wen2021pmp} formulates shape prediction as a point cloud deformation problem, and generates complete point clouds by moving input points to appropriate positions iteratively with a minimum moving distance.
Researchers have also introduced Transformer \cite{vaswani2017attention} for point cloud completion, like \cite{yu2021pointr,xiang2021snowflakenet}. PoinTr \cite{yu2021pointr} reformulates point cloud completion as a set-to-set translation problem, so that a transformer-based encoder-decoder architecture is naturally adopted. SnowflakeNet~\cite{xiang2021snowflakenet} utilizes transformer and point-wise feature deconvolutional modules to refine the initial point cloud iteratively.

Apart from the above supervised approaches, there are also a few studies concerning unpaired point cloud completion. AML \cite{stutz2018learning} introduces a weakly-supervised approach by learning the maximum likelihood. Pcl2Pcl \cite{chen2019unpaired} proposes a GAN framework to bridge the semantic gap between incomplete and complete shapes.
However, these methods only consider the single-side correspondence from the incomplete shapes to their complete shapes. In contrast, Cycle4Completion \cite{wen2021cycle4completion} enhances the completion performance by establishing the two-way geometric correspondence between complete shapes and incomplete shapes. Although achieving remarkable performance on point cloud completion, recent efforts generally use the mono-modality input. It is difficult to infer an accurate mapping from an incomplete point cloud with a large-scale incompleteness to its complete point cloud.

\subsection{Single-view Reconstruction}
Single-view reconstruction methods can be divided into point-based, voxel-based, mesh-based, and implicit field-based categories according to the representation of 3D data. Point-based methods~\cite{fan2017point,groueix2018papier,jiang2018gal} infer the point coordinates directly from images. PSGN~\cite{fan2017point} can reconstruct better shapes by connecting a 2D encoder and a 3D decoder. GAL~\cite{jiang2018gal} proposes a geometric adversarial loss to constrain the predicted point cloud. Voxel-based methods~\cite{choy20163d,xie2019pix2vox,xie2020pix2vox++} solve the reconstruction problem by using 2D and 3D CNNs. As for mesh-based methods, the pioneering work Pixel2Mesh~\cite{wang2018pixel2mesh} and Pixel2Mesh++~\cite{wen2019pixel2mesh++} fuse 2D features into mesh deformation. The follow-up methods~\cite{pan2019deep,li2020self} further extend this framework by improving deformation with topology modification and self-supervised learning. However, due to the intersection of meshes, the connection pattern hinders the generation of details. Recently, the community has focused heavily on deep implicit representations~\cite{park2019deepsdf,mescheder2019occupancy,mildenhall2021nerf,lin2020sdf,duggal2022topologically,mittal2022autosdf}, where 3D shapes are represented as implicit functions implemented by neural networks with possible image inputs. This representation has also been used to support other tasks, such as shape completion~\cite{mittal2022autosdf}, pairwise shape mating~\cite{chen2022neural}, and point cloud upsampling~\cite{zhao2022self}. Instead of directly inferring 3D shapes from single-view images, our approach assists the completion by combining different modalities at the feature level, where image features are used for both coarse completion and detailed refinement.

\subsection{Cross-modal Feature Fusion}
When it comes to data of ``cross-modality" (i.e., 3D point cloud and 2D image), the main difficulty is to fuse information provided by different modalities, which is a well-recognized challenge in various tasks such as 3D object detection, point cloud registration, and shape completion. 
ImVoteNet \cite{qi2020imvotenet} fuses 2D detection into a 3D pipeline with camera parameters and pixel depths. 
DeepI2P \cite{li2021deepi2p} utilizes the attention mechanism to weight image features by point clouds. 
Image2Point \cite{xu2021image2point} investigates the potential for transferability between single images and points. 
In the same problem setting, ViPC \cite{zhang2021view} also infers the missing points guided by an additional image. It directly maps a color image to the point cloud domain, explicitly aligns the generated rough geometry, and finally fine-tunes the coarse geometry. Although exploiting multi-modal information, ViPC heavily relies on the 3D reconstruction from a single image as well as requires an accurate alignment.
Different from ViPC, our CSDN formulates multi-modal fusion as a shape style transfer problem, and it disentangles different types of features with a dual-refinement module.

%% file: method.tex
\section{method}
\subsection{Overview}
We assume in point cloud completion with a cross-modal manner that 1) images supply both the global structure and local-detail information to facilitate the inference of a complete and geometry-rich shape, and 2) images should participate in the whole coarse-to-fine completion procedure. To this end, we design a Cross-modal Shape-transfer Dual-refinement Network
(CSDN). Following the encoder-decoder architecture to implement the cross-modal coarse-to-fine effort,  CSDN is designed to have three modules: feature extraction, shape fusion, and dual-refinement, as illustrated in Figure~\ref{fig:overview}.
Denote $P_{in}=\{p_1,...,p_N\}\subseteq\mathbb{R}^3$ with the size of $N$ to be an input partial point cloud, and $\mathcal{I}$ with the size of $H \times W$ to be the corresponding single-view color image. Our goal is to predict a point cloud $P_{out}$ from its dual-domain partial observations $P_{in}$ and $\mathcal{I}$, and $P_{out}$ represents the complete underlying shape of the object.

\subsection{Feature Extraction} 
CSDN embeds both the pair of a partial point cloud and a single-view image into their individual feature space.
Since we adopt a coarse-to-fine strategy (complete its overall shape first, and then complete its detail), we intend to extract global features from both \emph{$P_{in}$} and \emph{$I$} in this stage: 1) for \emph{$P_{in}$}, the simple yet effective PointNet \cite{qi2017pointnet} is leveraged to yield its global feature \emph{$F_p$} with the size of $1 \times C$; 2) for \emph{$I$}, a sub-network with seven convolutional layers is utilized to extract $7 \times 7 \times C$ feature maps and then obtain its global feature \emph{$F_I$} through average pooling.




\begin{figure}[hb]
  \centering
  \includegraphics[width=\linewidth]{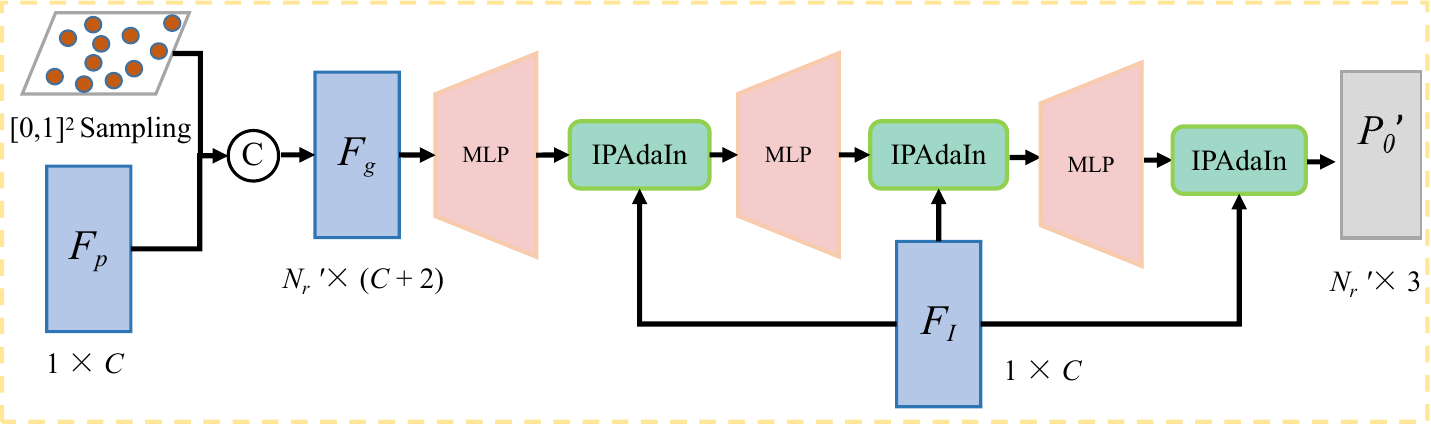}
  \caption{Illustration of the IPAdaIN layer. IPAdaIN normalizes features of the folding operation by global image features.}
  \label{fig:transfer}
\end{figure}
\subsection{Shape Fusion} 

We design the shape fusion module to fuse the features of different modalities, which outputs a coarse but complete set of shape points. 
By proposing an image and point cloud co-supported AdaIN \cite{huang2017arbitrary} (IPAdaIN for short), we transfer the intrinsic geometric style from the image to the point cloud. 

This module takes \emph{$F_p$} and \emph{$F_I$} as input and outputs a coarse point cloud \emph{$P_0$} (see Figure \ref{fig:transfer}). Specifically, similar to FoldingNet~\cite{yang2018foldingnet}, we first sample $N_r'$ 2D points from a unit square $[0, 1]^2$. Each of them is appended with the global point cloud feature $F_p$. Then, the concatenated feature is combined with the global image feature \emph{$F_I$} to fold the 2D grids to a point surface, by the IPAdaIN layer. Note that we repeat (but without sharing the network parameters) the above operations $M$ times to reconstruct $M$ surfaces, similar to \cite{groueix2018papier,liu2020morphing}. Finally, all of them are concatenated to synthesize a complete complex point cloud {$P_0$} with $N_r$ points, i.e., $N_r$ = $M \times N_r'$. $M$ is a hyperparameter fixed to 4 in our experiments. Also, we can change $K$ and $N_r'$ to adapt the proposed method to the point clouds of different resolutions.

\textbf{IPAdaIN.} Inspired by StyleGAN~\cite{karras2019style} that employs AdaIN for the style transfer task and SpareNet~\cite{xie2021style} that generates points with a similar batch normalization operation, we reformulate the ``cross-modality" fusion as a shape style transfer problem.
The core idea is that the affine parameters in Instance Normalization generated by different image features can change the point-wise feature statistics to varying values and thus normalize the output point cloud to different shapes. Meanwhile, the global features learned by convolutional neural networks can describe the overall shape characteristics.
Accordingly, we propose IPAdaIN, a new solution that is built based on the instance normalization layer to better capture the image style. Specifically, we feed the concatenated feature to multiple MLPs, where each MLP is followed by an IPAdaIN layer and generates a new point-wise feature $F_{in}$. By denoting $F_{in}\subseteq\mathbb{R}^{B\times N\times C}$ (B stands for the batch size during training), the normalization process is formulated as
\begin{equation} \label{eqn1}
  \begin{split}
  IPAdaIN(F_{in},F_I) &= \gamma(\frac{F_{in}-\mu(F_{in})}{\sigma(F_{in})}) + \beta   \\
  \gamma &= L_a(F_I) \\
  \beta &= L_b(F_I)
  \end{split},
\end{equation}
where $\gamma\subseteq\mathbb{R}^{B\times 1\times C}$ and $\beta\subseteq\mathbb{R}^{B\times 1\times C}$ are the affine parameters computed by non-linear mappings implemented by two MLPs.
$\mu(F_{in})\subseteq\mathbb{R}^{B\times 1\times C}$ and $\sigma(F_{in})\subseteq\mathbb{R}^{B\times 1\times C}$ are the means and standard deviations of the channel-wise activation of $F_{in}$, respectively
\begin{equation} \label{eqn2}
  \begin{split}
  \mu_{bc}(F_{in}) &= \frac{1}{N}\sum_{n=1}^{N}{f_{bcn}}\\
  \sigma_{bc}(F_{in}) &=
  \sqrt{\frac{1}{N}\sum_{n=1}^{N}{(f_{bcn} - \mu_{bc}(F_{in}))^2} + \epsilon}
  \end{split}.
\end{equation}
Unlike AdaIN, IPAdaIN guides shape folding with the learned shape characteristics, which can be interpreted as a way to reconstruct a 3D shape from its style contained in the image. 
Considering that different feature channels of an image can identify certain shape characteristics, the generated affine parameters $\gamma$ and $\beta$ will have high activation in varying channels according to the 3D shapes. Then the point-wise features change in every channel according to Eq.~\ref{eqn1}. The fused features are then decoded back to the spatial space by the 3D decoder of the folding operation.
\begin{figure}[h]
  \centering
  \includegraphics[width=\linewidth]{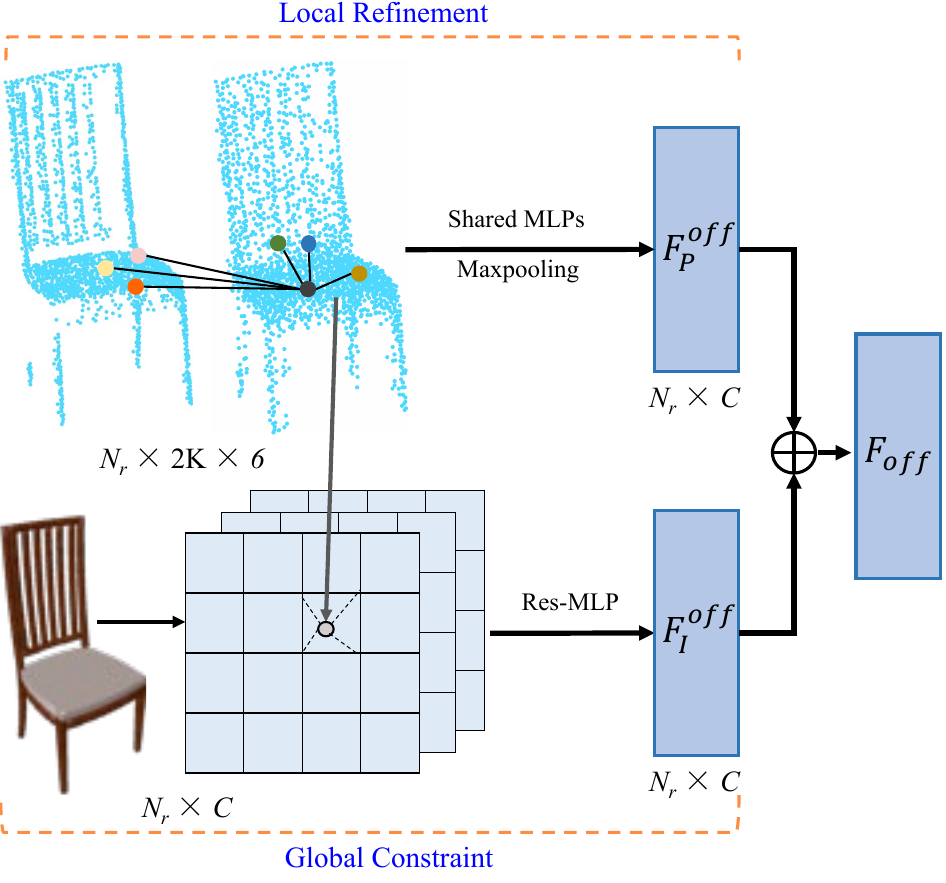}
  \caption{Illustration of the dual-refinement module. The upper path denotes Local Refinement and the lower one denotes Global Constraint. Each sub-network generates an offset feature, and they are added to yield the feature $F_{off}$. }
  \label{fig:localgraph}
\end{figure}
\subsection{Dual-Refinement}
\label{sec:dual}
The dual-refinement module aims to generate a set of coordinate offsets $\Delta C$ with the dimension of $N_r \times 3$ to fine-tune the coarse point cloud \emph{$P_0$}. The whole module contains two units named Local Refinement and Global Constraint, respectively. The local refinement finds the relation between points in \emph{$P_0$} and \emph{$P_{in}$} and generates one set of offset features, as shown in Figure \ref{fig:localgraph}.
To refine $P_0$ without being corrupted by missing parts, the global constraint exploits image features to generate another set of offset features. Finally, the combined feature is fed into the Offset Regression unit that is built with an MLP to obtain the set of offsets $\Delta C$. Thus, the two units of local refinement and global constraint complement each other to capture both the global structures and local details of the completed object. 

\subsubsection{Local Refinement}
It is a very common situation that some missing high-frequency details already appear in the partial point cloud. Thus an intuition is to exploit local structure information with the aid of the partial point cloud. 
The local refinement unit is to integrate the structure information of the input partial shape with the coarse point cloud itself. Since a point cloud can be naturally regarded as a graph-like structure, we build this unit with graph convolution, given its effectiveness in point cloud analysis \cite{iccv/ZhouFFW0L21}. 

A directed graph $\mathcal{G}(\mathcal{V},\mathcal{E})$ is  established from the partial point cloud $P_{in}$ and the coarse point cloud $P_0$, where $\mathcal{V} = \{1,...,N+N_r\}$ represents the set of nodes, and $\mathcal{E} \subseteq \mathcal{V} \times \mathcal{V}$ is the set of edges. Observing that certain areas in $P_{in}$ have much higher quality than those in $P_{0}$, like the chair back and the three complete legs in Figure \ref{fig:localgraph}, we build $\mathcal{E}$ encoding the structure information from both \emph{$P_0$} and \emph{$P_{in}$}. Hence, for each point in $P_0$, its $k$-nearest neighbours are respectively searched in \emph{$P_{in}$} and \emph{$P_0$} to build a local graph which has 2k vertices (see Figure \ref{fig:localgraph}). By denoting the central point in the graph convolution as \emph{$p_i$} and the neighborhood set as $\mathcal{N}(i) = \{j | (i,j) \in \mathcal{E}\}$, we can formulate \emph{edge features} as $e_{ij}=\sigma([p_i,p_j-p_i])$, where $\sigma$ is a non-linear function implemented by Shared-MLP and $[\cdot,\cdot]$ is a concatenation operation. Finally, each point's feature is aggregated by
\begin{equation}
    f_{i}=\max \limits_{j\in \mathcal{N}(i)}e_{ij},
\end{equation}
and
\begin{equation}
    F_P^{off} = \{f_{i}\}_{i=1}^{N_r},
\end{equation}
where max denotes a channel-wise max-pooling function and $f_{i}$ is the feature of the $i$-th point.

\subsubsection{Global Constraint}
We query $k$-nearest neighbors from the partial point cloud \emph{$P_{in}$} for each point in $P_0$. Inevitably, the points in the generated missing areas cannot contain a reliable local graph similar to the corresponding local shape in the ground-truth point cloud. Meanwhile, RGB images usually contain underlying 3D-aware shape properties with high confidence (e.g., boundaries, textures, and local connectivity)~\cite{cgf/NanSC14}. To this end, we propose the global constraint unit to fix $F_P^{off}$ with learned image features. The unit first projects 3D points into the last four feature maps (from 2D encoders) by the camera parameters and pooling them from four nearby pixels using bilinear interpolation (see the bottom of Figure \ref{fig:localgraph}). The obtained point-wise feature is then fed into a residual MLP block to obtain the offset feature \emph{$F_{I}^{off}$}. Finally, we add \emph{$F_{I}^{off}$} and \emph{$F_{P}^{off}$} to obtain \emph{$F_{off}$}.

\begin{figure*}[ht] 
  \centering
  \includegraphics[width=0.96\textwidth]{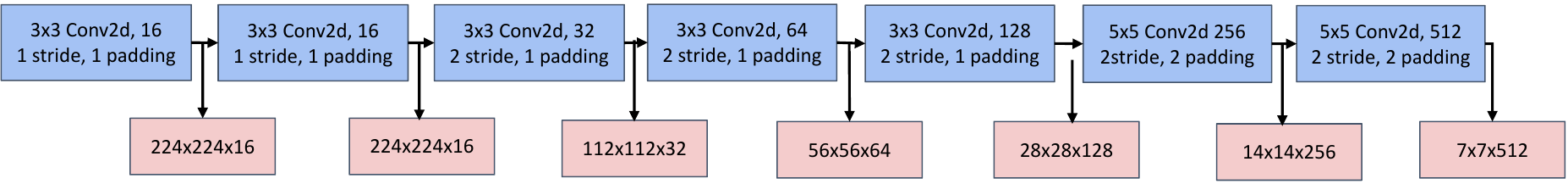}
  \caption{The architecture of our 2D Encoder.}
  \label{fig:cnn}
\end{figure*}

\subsection{Loss Function}
The loss function measures the difference between  \emph{$P_{out}$} and the ground truth \emph{$P_{gt}$}. We utilize the Chamfer distance (CD) as the loss function. 
To achieve the coarse-to-fine generation process, we regularize the training by calculating the loss function as
\begin{equation}
    \mathcal{L}= \mathcal{L}_{CD}(P_0,P_{gt})+\alpha\mathcal{L}_{CD}(P_{out},P_{gt}),
\end{equation}
where $\mathcal{L}_{CD}$ is defined as
\begin{equation}
    \mathcal{L}_{CD}(X,Y)=\frac{1}{\lvert X \rvert}\sum_{x \in X}\min_{y \in Y} \lvert \lvert x-y \rvert \rvert + \frac{1}{\lvert Y \rvert}\sum_{y \in Y}\min_{x \in X} \lvert \lvert y-x \rvert \rvert.
\end{equation}
In the implementation, we increase $\alpha$ from 0.01 to 2.0 during the first 30k iterations since the initial point set \emph{$P_0$} is less accurate. 
\begin{table*}
    \tiny
    \renewcommand\arraystretch{1.2}
    \centering
    \caption{Quantitative results on known categories of ShapNet-ViPC using CD with 2,048 points. The best is highlighted in bold.}
    \label{tab:tab1}
    \normalsize
    \begin{tabular}{c|c|c|c|c|c|c|c|c|c}
    \hline
    \multirow{2}{*}{Methods} & \multicolumn{9}{c}{Mean CD per point $\times 10^{-3}$ (lower is better)} \cr\cline{2-10}
    &Avg & Airplane & Cabinet & Car & Chair & Lamp & Sofa & Table & Watercraft \cr
    \hline
    \hline
              AtlasNet \cite{groueix2018papier}  & 6.062 & 5.032 & 6.414 & 4.868 & 8.161 & 7.182 & 6.023 & 6.561 & 4.261 \\
              \hline
              FoldingNet \cite{yang2018foldingnet}  & 6.271 & 5.242 & 6.958 & 5.307 & 8.823 & 6.504 & 6.368 & 7.080 & 3.882 \\
              \hline
              PCN \cite{yuan2018pcn}  & 5.619 & 4.246 & 6.409 & 4.840 & 7.441 & 6.331 & 5.668 & 6.508 & 3.510 \\
              \hline
              TopNet \cite{tchapmi2019topnet}  &4.976 & 3.710 & 5.629 & 4.530 & 6.391 & 5.547 & 5.281 & 5.381 & 3.35 \\
              \hline
              PF-Net \cite{huang2020pf}  & 3.873 & 2.515 & 4.453 & 3.602 & 4.478 & 5.185 & 4.113 & 3.838 & 2.871 \\
              \hline
              MSN \cite{liu2020morphing}  & 3.793 & 2.038 & 5.06 & 4.322 & 4.135 & 4.247 & 4.183 & 3.976 & 2.379 \\
              \hline
              GRNet \cite{xie2020grnet}  & 3.171 & 1.916 & 4.468 & 3.915 & 3.402 & 3.034 & 3.872 & 3.071 & 2.160 \\
              \hline
              PoinTr \cite{yu2021pointr}  & 2.851 & 1.686 & 4.001 & 3.203 & 3.111 & 2.928 & 3.507 & 2.845 & 1.737 \\
              \hline
              ViPC \cite{zhang2021view}  & 3.308 & 1.760 & 4.558 & 3.138 & \textbf{2.476} & 2.867 & 4.481 & 4.99 & 2.197 \\
              \hline
              PointAttN \cite{wang2022pointattn}  & 2.853 & 1.613 & 3.969 & 3.257 & 3.157 & 3.058 & 3.406 & 2.787 & 1.872 \\
              \hline
              SDT \cite{zhang2022point}  & 4.246 & 3.166 & 4.807 & 3.607 & 5.056 & 6.101 & 4.525 & 3.995 & 2.856 \\
              \hline
              Seedformer \cite{zhou2022seedformer}  & 2.902 & 1.716 & 4.049 & 3.392 & 3.151 & 3.226 & 3.603 & 2.803 & \textbf{1.679} \\
              \hline
              Ours & \textbf{2.570} & \textbf{1.251} & \textbf{3.670} & \textbf{2.977} & 2.835 & \textbf{2.554} & \textbf{3.240} & \textbf{2.575} & 1.742 \\
              \hline
    \hline
    \end{tabular}
\end{table*}
\begin{table*}[!t]
\tiny
    \renewcommand\arraystretch{1.2}
        \centering
        \caption{Quantitative results on known categories of ShapNet-ViPC using F-Score with 2,048 points. The best is highlighted in bold.}
        \label{tab:tab2}
        \normalsize
        \begin{tabular}{c|c|c|c|c|c|c|c|c|c}
        \hline
        \multirow{2}{*}{Methods}& 
        \multicolumn{9}{c}{F-Score@0.001 (higher is better)} \cr\cline{2-10}
        & Avg & Airplane & Cabinet & Car & Chair & Lamp & Sofa & Table & Watercraft \cr
        \hline
        \hline
                  AtlasNet \cite{groueix2018papier} & 0.410 & 0.509 & 0.304 & 0.379 & 0.326 & 0.426 & 0.318 & 0.469 & 0.551 \\
                  \hline
                  FoldingNet \cite{yang2018foldingnet} & 0.331 & 0.432 & 0.237 & 0.300 & 0.204 & 0.360 & 0.249 & 0.351 & 0.518 \\
                  \hline
                  PCN \cite{yuan2018pcn} & 0.407 & 0.578 & 0.27 & 0.331 & 0.323 & 0.456 & 0.293 & 0.431 & 0.577 \\
                  \hline
                  TopNet \cite{tchapmi2019topnet}  &0.467 & 0.593 & 0.358 & 0.405 & 0.388 & 0.491 & 0.361 & 0.528 & 0.615 \\
                  \hline
                  PF-Net \cite{huang2020pf}  & 0.551 & 0.718 & 0.399 & 0.453 & 0.489 & 0.559 & 0.409 & 0.614 & 0.656 \\
                  \hline
                  MSN \cite{liu2020morphing} & 0.578 & 0.798 & 0.378 & 0.380 & 0.562 & 0.652 & 0.410 & 0.615 & 0.708 \\
                  \hline
                  GRNet \cite{xie2020grnet} & 0.601 & 0.767 & 0.426 & 0.446 & 0.575 & 0.694 & 0.450 & 0.639 & 0.704 \\
                  \hline
                  PoinTr \cite{yu2021pointr} & 0.683 & 0.842 & 0.516 & 0.545 & 0.662 & 0.742 & 0.547 & 0.723 & 0.780 \\
                  \hline
                  ViPC \cite{zhang2021view}  & 0.591 & 0.803 & 0.451 & 0.5118 & 0.529 & 0.706 & 0.434 & 0.594 & 0.73 \\
                  \hline
                  PointAttN \cite{wang2022pointattn}  & 0.662 & 0.841 & 0.483 & 0.515 & 0.638 & 0.729 & 0.512 & 0.699 & 0.774 \\
                  \hline
                  SDT \cite{zhang2022point}  & 0.473 & 0.636 & 0.291 & 0.363 & 0.398 & 0.442 & 0.307 & 0.574 & 0.602 \\
                  \hline
                  Seedformer \cite{zhou2022seedformer}  & 0.688 & 0.835 & \textbf{0.551} & 0.544 & 0.668 & \textbf{0.777} & 0.555 & 0.716 & \textbf{0.786} \\
                  \hline
                  Ours & \textbf{0.695} & \textbf{0.862} & 0.548 & \textbf{0.560} & \textbf{0.669} & 0.761 & \textbf{0.557} & \textbf{0.729} & 0.782 \\
                  \hline
        \hline
        \end{tabular}
    \end{table*}

%% file: experiment.tex
\section{experiment}
\begin{figure*}[ht]
  \centering
  \includegraphics[width=\textwidth]{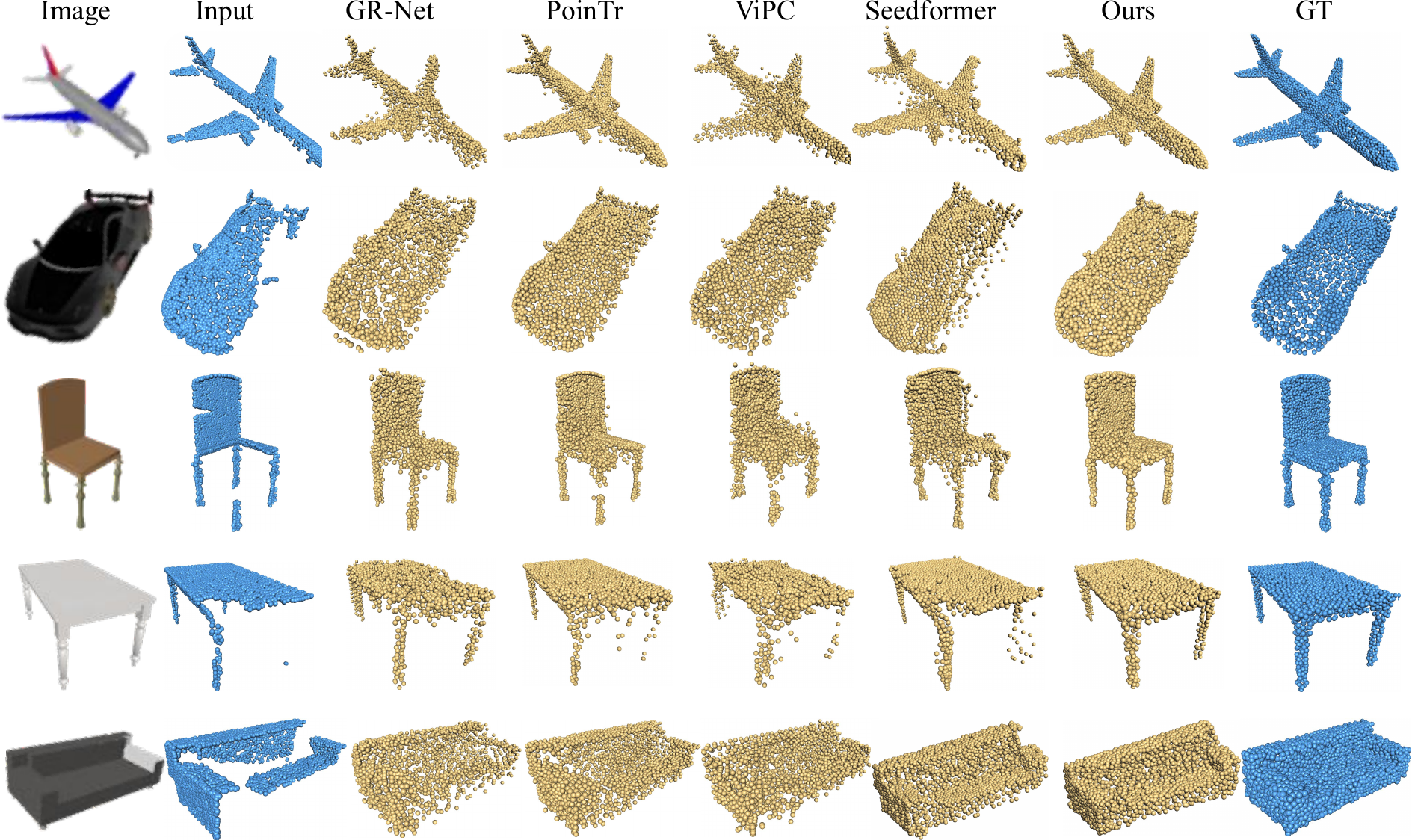}
  \caption{Visual comparisons of recent point cloud completion methods~\cite{xie2020grnet,yu2021pointr,zhang2021view,zhou2022seedformer} on ShapeNet-ViPC~\cite{zhang2021view}. CSDN produces the most complete and detailed structures compared to its competitors.}
  \label{fig:visualcomp}
\end{figure*} 
\begin{figure*}[h]
  \centering
  \includegraphics[width=\textwidth]{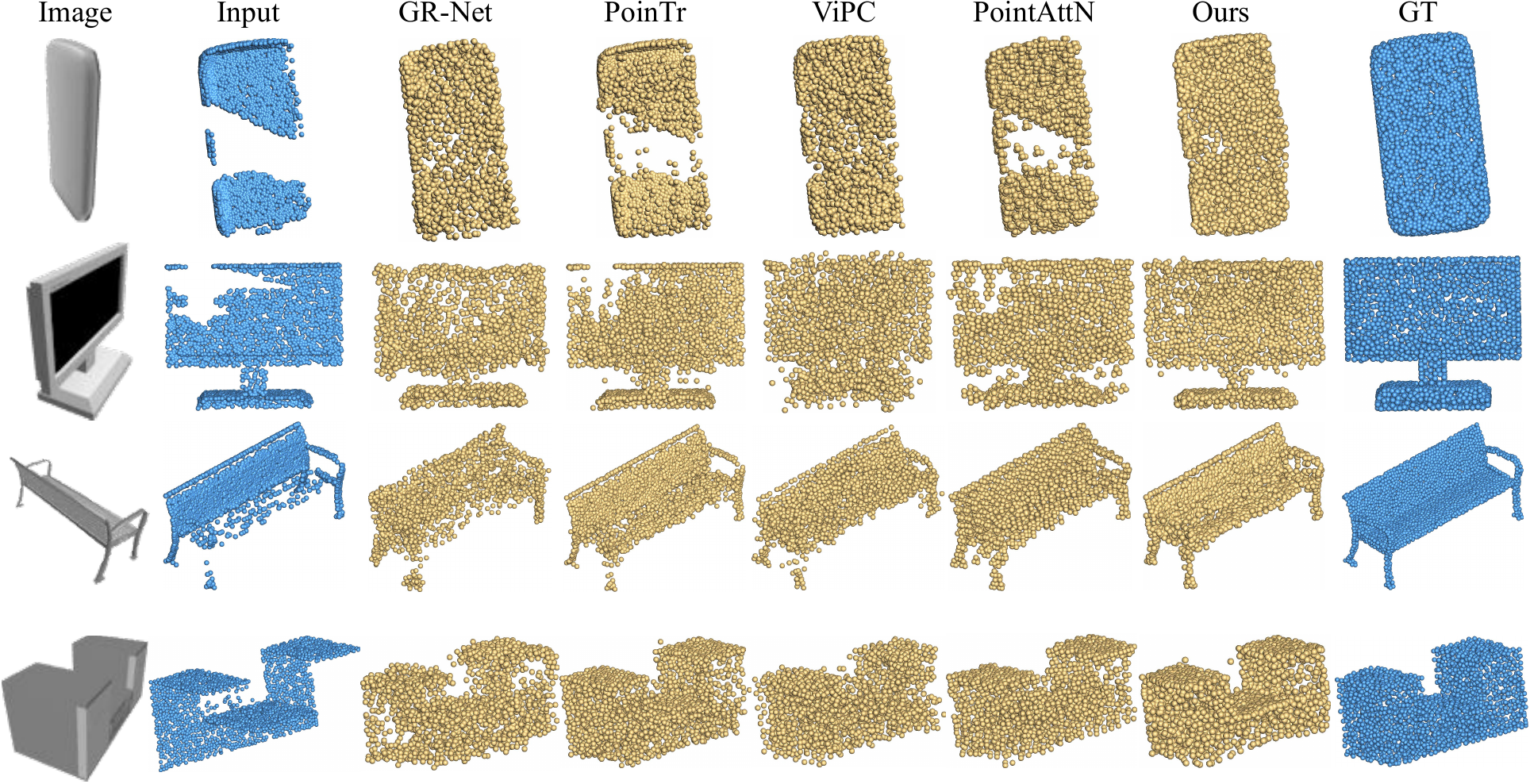}
  \caption{Visual comparisons of recent point cloud completion methods~\cite{xie2020grnet,yu2021pointr,zhang2021view,wang2022pointattn} on unseen categories of ShapeNet-ViPC~\cite{zhang2021view}. CSDN produces the most complete and detailed structures compared to its competitors.}
  \label{fig:visnovel}
\end{figure*}  
\subsection{Datasets}
\subsubsection{Training Dataset}
We use the benchmark dataset ShapeNet-ViPC \cite{zhang2021view}, which is derived from ShapeNet \cite{chang2015shapenet}. 
It contains 13 categories and 38,328 objects, including airplane, bench, cabinet, car, chair, monitor, lamp, speaker, firearm, sofa, table, cellphone, and watercraft. The complete ground-truth point cloud is generated by uniformly sampling 2048 points on the mesh surface from ShapeNet \cite{chang2015shapenet}. Each object has a complete point cloud, 24 rendered images under 24 viewpoints, and 24 corresponding incomplete point clouds. The incomplete point cloud is generated from the corresponding viewpoint (with occlusion) and also contains 2048 points. The image data is rendered from 24 viewpoints as ShapeNetRendering in 3D-R2N2 \cite{choy20163d}. During training, the image viewpoint is randomly chosen for each training data pair and the point cloud will be aligned with the chosen image. For a fair comparison, we use the same training setting as in ViPC \cite{zhang2021view}, i.e., 80\% of the eight categories for training.

\subsubsection{Test Dataset}
The test dataset from ShapeNet-ViPC \cite{zhang2021view} consists of two parts: one contains the remaining 20\% of the 8 categories of objects for training; another one contains 4 categories that are not used during training. For each of the new categories, we randomly choose 200 pairs of point clouds and images. For a fair comparison, we follow the same setting as \cite{zhang2021view}, where images are randomly selected from the 24 viewpoints. Notably, the ShapeNet-ViPC dataset produces missing shapes by various kinds of occlusions (not limited to self-occlusion). Therefore, images from different viewpoints do not necessarily depict the missing parts. We test our approach when images and partial point clouds are produced under the same viewpoints. The results are reported in Section~\ref{viewEXP}.

Besides, we test some methods on partial point clouds from real-world LiDAR scans, the KITTI dataset~\cite{geiger2013vision} and RGB-D scans, the Scannet dataset~\cite{dai2017scannet}. Following the setting of \cite{yuan2018pcn}, we extract point clouds within the corresponding bounding boxes and then select point clouds having more than 2048 points.
\subsection{Implementation Details}
\label{detail}
In our implementation, the number of point surfaces in Shape Fusion is $k=4$ and each surface contains 512 points. The feature dimension is set to 1024, i.e., C = 1024. Thus, the generated coarse point cloud has 2048 points. For IPAdaIN, we set three layers for every surface. In the Local Refinement unit, the number of nearest neighbors is 16. In the Global Constraint unit, the size of feature maps used for projection are $56\times56$, $28\times28$, $14\times14$, and $7\times7$.

For the 3D encoder, we set a five-layer MLP. The feature dimensions are 64, 64, 64, 128, and 1024, respectively. For the 2D encoder, its architecture is shown in Figure \ref{fig:cnn}. 
Note that we reuse the last four feature maps in the global constraint unit and their dimensions are $56\times56$, $28\times28$, $14\times14$, and $7\times7$, respectively.
In the local refinement unit, the feature \emph{$F_{P}^{off}$} is produced by a three-layer MLP whose output dimensions are 32, 128, and 512, respectively. Res-MLP in the global constraint unit is implemented by a Con1d ResNet block \cite{he2016deep}. The final offset regression is implemented by four Conv1d layers whose output dimensions are 256, 128, 32, and 3, respectively. The coordinate offset is calculated through the \emph{tanh} activation function.

For quantitative evaluation, we use both the Chamfer Distance (CD) and F-Score as evaluation metrics. The whole network is trained end-to-end with a learning rate of $5 \times 10^{-5}$ for 50 epochs and the Adam optimizer. The learning rate also decayed by 0.1 for every 10 epochs. All the networks are implemented using PyTorch and trained on an NVIDIA RTX 3090 GPU.

\subsection{Result on ShapeNet-ViPC}
\label{subsec:comparison}
We compare our CSDN with ten state-of-the-art point cloud completion methods. They are Point-based, i.e., AtlasNet \cite{groueix2018papier}, PCN \cite{yuan2018pcn}, MSN \cite{liu2020morphing}, Transformer-based, i.e., PoinTr \cite{yu2021pointr}, Seedformer~\cite{zhou2022seedformer}, SDT~\cite{zhang2022point}, Folding-based, i.e., FoldingNet \cite{yang2018foldingnet}, TopNet \cite{tchapmi2019topnet}, GAN-based, i.e., PF-Net \cite{huang2020pf},  Convolution-based, i.e., GRNet \cite{xie2020grnet}, and Cross-modality-based, i.e., ViPC \cite{zhang2021view}. For AtlasNet, FoldingNet, PCN, and TopNet, we directly take the quantitative statistics from \cite{zhang2021view}. For the other methods, we retrain their models using the released codes on ShapeNet-ViPC. 
For a fair comparison, we uniformly downsample the results of PoinTr to 2048 points, the same as the output of the other methods. Among these methods, ViPC also exploits an additional image, while the others take only a partial point cloud as input.

\subsubsection{Qualitative Results on Known Categories}
Figure \ref{fig:visualcomp} shows the point clouds completed by the proposed CSDN and its competitors. 
For some cases of symmetric structures and small missing regions (e.g., plane and car), most methods behave well and generate complete shapes similar to the ground truths. Our CSDN recovers more clear overall shapes and more detailed structures than Seedformer, PoinTr, and ViPC, while GR-Net generates noisy points. For those challenging cases of complex shapes and large missing regions, like the table case, our CSDN generates the complete desktop and leg while the other methods only predict a coarse result. Compared to ViPC, 
our CSDN is better to recover latent geometric structures (e.g., holes in the desktop and the back of the chair).

\subsubsection{Quantitative Results on Known Categories}
We calculate both CD and F-Score on the 2,048 points of each object, as reported in Table \ref{tab:tab1} and Table \ref{tab:tab2}. It can be observed that our CSDN achieves almost the best results among all the competitors. It is worth noting that CSDN reduces the average CD value by 0.281 compared to the result of the second-ranked PoinTr \cite{yu2021pointr}. Also, CSDN outperforms ViPC \cite{zhang2021view} by a large margin of 22.3\%. Meanwhile, CSDN achieves the best results in all categories in terms of F-Score. 
    
\subsubsection{Results on Novel Categories}
We test different methods on four novel categories which are not used for training, to evaluate the category-agnostic ability. The quantitative results are reported in Table~\ref{tab:tab3} and Table~\ref{tab:tab4}. PoinTr~\cite{yu2021pointr} and PointAttN~\cite{wang2022pointattn} achieve competitive results similar to their results on known categories. ViPC~\cite{zhang2021view} struggles to generalize to new shapes, while our CSDN generalizes well to novel shapes that have not been seen during training.

We also present a visual comparison with other methods, including two Transformer-based methods PoinTr~\cite{yu2021pointr} and PointAttN~\cite{wang2022pointattn}, one Convolution-based method GR-Net~\cite{xie2020grnet}, and one Cross-modality-based method ViPC~\cite{zhang2021view}, as shown in Figure~\ref{fig:visnovel}. We can observe that the two Transformer-based methods cannot well recover the missing shapes for previously unseen categories, although they achieve competitive performance on quantitative results. Under the guidance of images, CSDN and ViPC both succeed in inferring the missing parts, and CSDN produces better details and fewer noisy points. This comparison further demonstrates that our method successfully exploits the complementary information provided by the images.

\begin{table}
\tiny
    \caption{Quantitative results on the Novel categories of ShapNet-ViPC using CD with 2,048 points. The best is highlighted in bold.}
    \renewcommand\arraystretch{1.2}
        \centering
        \label{tab:tab3}
        \footnotesize
        \normalsize
        \begin{tabular}{c|c|c|c|c|c}
        \hline
        \multirow{2}{*}{Methods}& 
        \multicolumn{5}{c}{Mean Chamfer Distance per point $\times 10^{-3}$} \cr\cline{2-6}
        & Avg & Bench & Monitor & Speaker & Phone \cr
        \hline
        \hline
                  PF-Net \cite{huang2020pf} & 5.011 & 3.684 & 5.304 & 7.663 & 3.392 \\
                  \hline
                  MSN \cite{liu2020morphing} & 4.684 & 2.613 & 4.818 & 8.259 & 3.047 \\
                  \hline
                  GRNet \cite{xie2020grnet} & 4.096 & 2.367 & 4.102 & 6.493 & 3.422 \\
                  \hline
                  PoinTr \cite{yu2021pointr} & 3.755 & 1.976 & 4.084 & 5.913 & 3.049 \\
                  \hline
                  ViPC \cite{zhang2021view} & 4.601 & 3.091 & 4.419 & 7.674 & 3.219 \\
                  \hline
                  PointAttN \cite{wang2022pointattn} & 3.674 & 2.135 & \textbf{3.741} & 5.973 & \textbf{2.848} \\
                  \hline
                  SDT \cite{zhang2022point} & 6.001 & 4.096 & 6.222 & 9.499 & 4.189 \\
                  \hline
                  Ours & \textbf{3.656} & \textbf{1.834} & 4.115 & \textbf{5.690} & 2.985 \\
                  \hline
        \hline
        \end{tabular}
    \end{table}
\begin{table}
\tiny
    \renewcommand\arraystretch{1.2}
        \centering
        \caption{Quantitative results on the Novel categories of ShapNet-ViPC using F-Score with 2,048 points. The best is highlighted in bold.}
        \label{tab:tab4}
        \footnotesize
        \normalsize
        \begin{tabular}{c|c|c|c|c|c}
        \hline
        \multirow{2}{*}{Methods}& 
        \multicolumn{5}{c}{F-Score@0.001} \cr\cline{2-6}
        & Avg & Bench & Monitor & Speaker & Phone \cr
        \hline
        \hline
                  PF-Net \cite{huang2020pf} & 0.468 & 0.584 & 0.433 & 0.319 & 0.534 \\
                  \hline
                  MSN \cite{liu2020morphing} & 0.533 & 0.706 & 0.527 & 0.291 & 0.607 \\
                  \hline
                  GRNet \cite{liu2020morphing} & 0.548 & 0.711 & 0.537 & 0.376 & 0.569 \\
                  \hline
                  PoinTr \cite{yu2021pointr} & 0.619 & 0.797 & \textbf{0.599} & 0.454 & 0.627 \\
                  \hline
                  ViPC \cite{zhang2021view} & 0.498 & 0.654 & 0.491 & 0.313 & 0.535 \\
                  \hline
                  PointAttN \cite{wang2022pointattn} & 0.605 & 0.764 & 0.591 & 0.428 & 0.637 \\
                  \hline
                  SDT \cite{zhang2022point} & 0.327 & 0.479 & 0.268 & 0.197 & 0.362 \\
                  \hline
                  Ours & \textbf{0.631} & \textbf{0.798} & 0.598 & \textbf{0.485} & \textbf{0.644} \\
                  \hline
        \hline
        \end{tabular}
\end{table}

\subsection{Evaluation on Real-world Scans}
\begin{figure}[h]
  \centering
  \includegraphics[width=\linewidth]{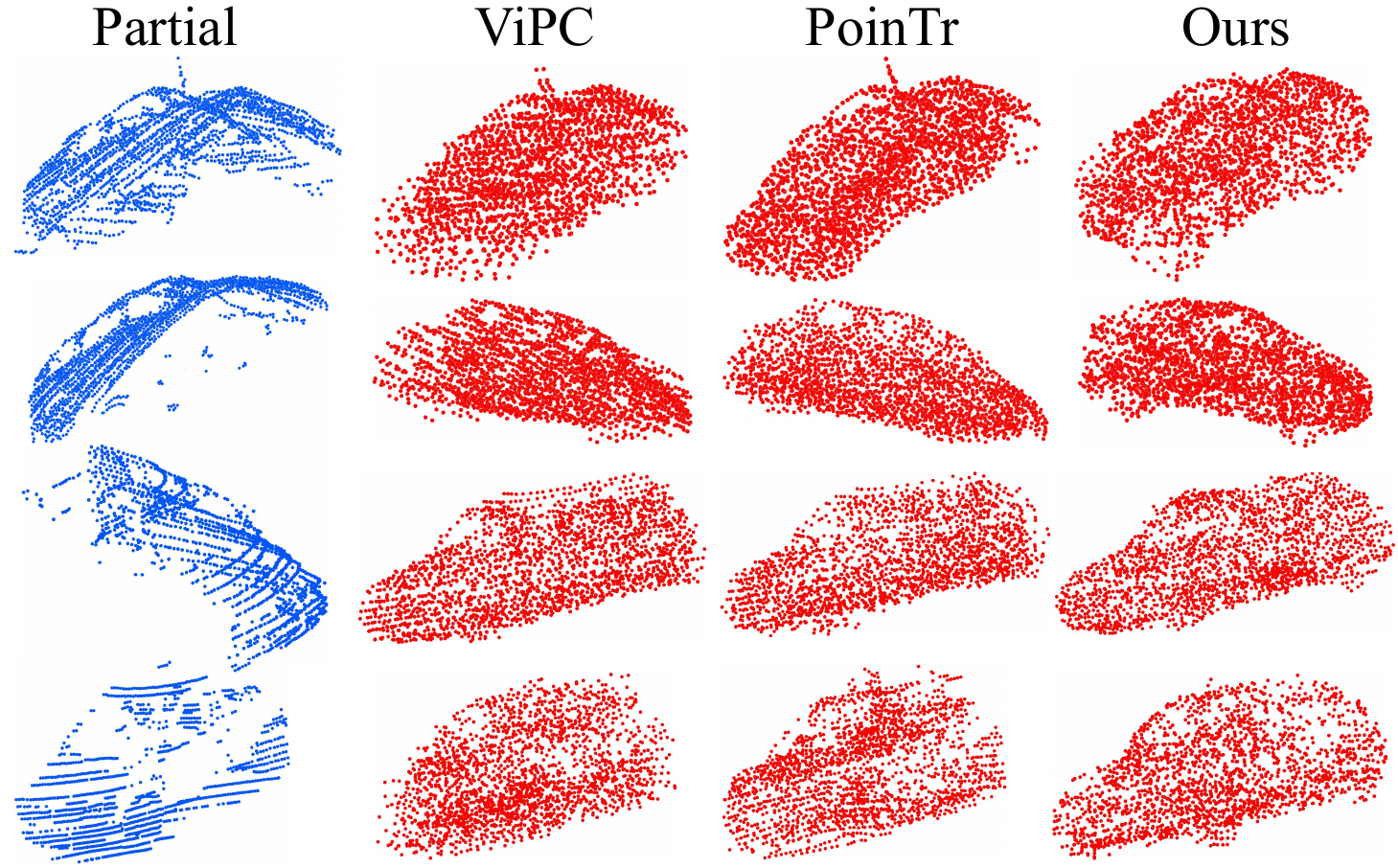}
  \caption{Visual comparisons with two point cloud completion methods~\cite{zhang2021view,yu2021pointr} on the real-scanned point clouds (from KITTI~\cite{geiger2013vision}). CSDN produces the most complete and detailed structures.}
  \label{fig:kitti}
\end{figure}

\begin{figure}[ht]
  \centering
  \includegraphics[width=\linewidth]{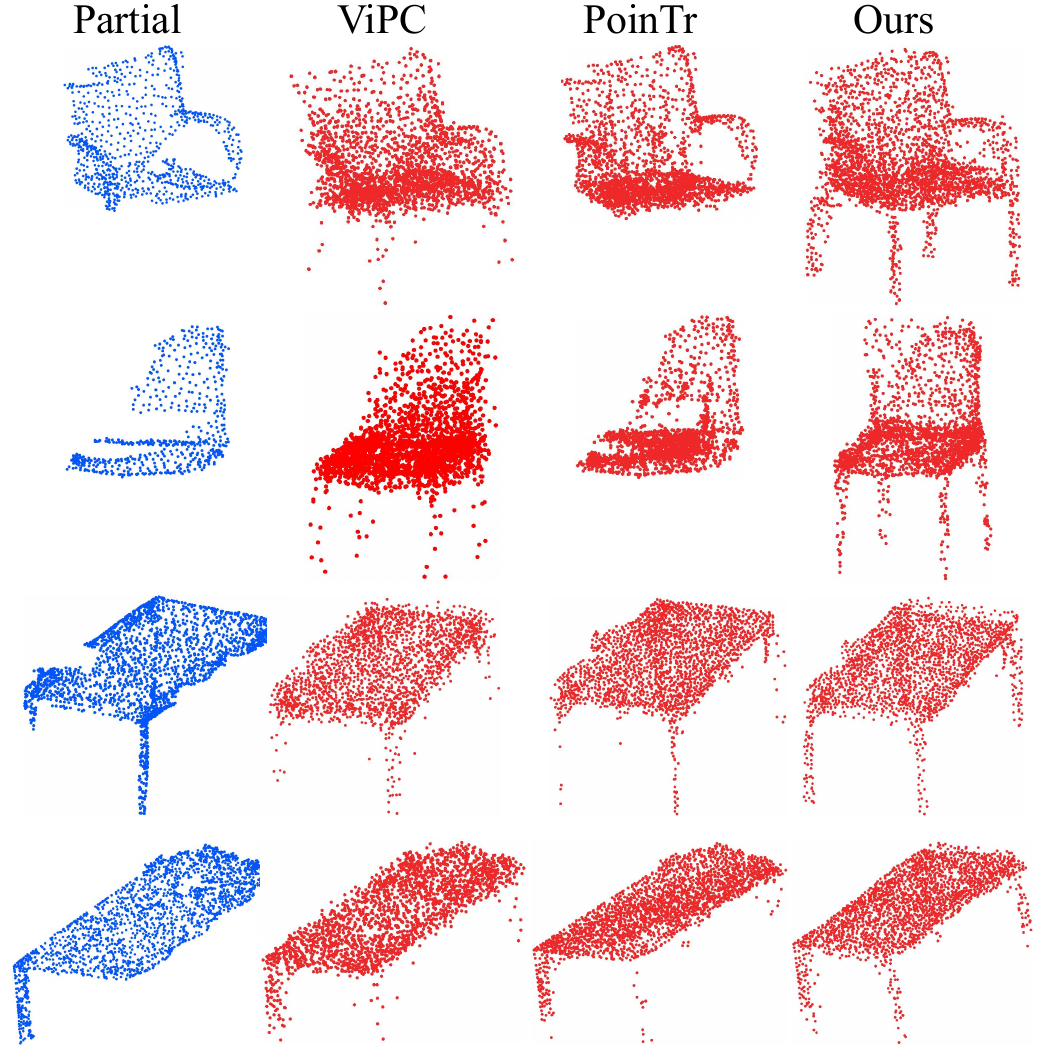}
  \caption{Visual comparisons with two point cloud completion methods~\cite{zhang2021view,yu2021pointr} on the real-scanned point clouds from Scannet~\cite{dai2017scannet}. CSDN produces the most complete and detailed structures.}
  \label{fig:scannet}
\end{figure}

We train our CSDN and two other methods on the car category of ShapeNet-ViPC to evaluate their performance on real-world scans. 
Ideally, a point cloud completion network trained on a cross-modal dataset should produce satisfactory results, since the cars and their corresponding real-world images cropped with 2D bounding boxes are provided. As reported in \cite{zhang2021view}, the real-to-synthetic domain gap led by rendered images in the training dataset makes ViPC not applicable to real-world point cloud completion. Thus we replace the input image with the rendered car image in ShapeNet-ViPC for ViPC~\cite{zhang2021view} and our method.  We also evaluate our method on the chairs and tables from the Scannet dataset~\cite{dai2017scannet}. Since the objects in Scannet are usually more complete and similar to those in the training datasets, we directly use the trained model in Section~\ref{subsec:comparison} without any fine-tuning. Figure~\ref{fig:kitti} shows that ViPC still cannot handle the domain gap problem and PoinTr generates coarse results due to the low resolution of the training dataset. In Figure~\ref{fig:scannet}, PoinTr fails to complete the missing parts, and ViPC cannot preserve the original inputs. In contrast, CSDN is capable of generating more plausible results on real-world scans.


\subsection{Ablation Study}
We first remove and change the main components  ablate CSDN. The ablation variants can be classified as ablation on Shape Fusion and Dual Refinement. Then, we ablate on the input modality and analyze the contribution of each modality.

\begin{figure*}[ht] 
  \centering
  \includegraphics[width=\textwidth]{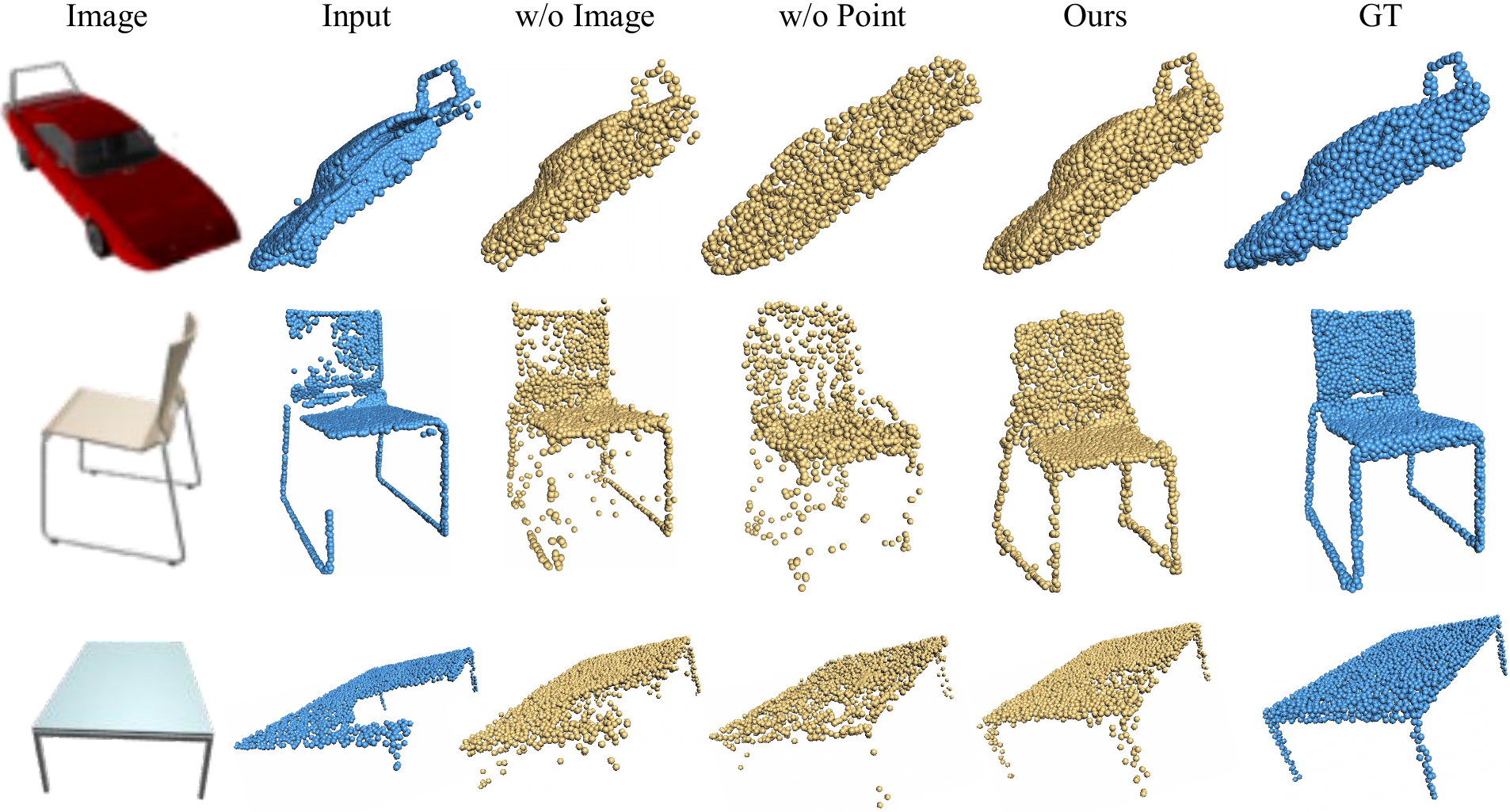}
  \caption{Visual comparisons between CSDN and its variants when the input images cannot depict the missing parts.}
  \label{fig:vis112}
\end{figure*}
\begin{figure*}[ht] 
  \centering
  \includegraphics[width=\textwidth]{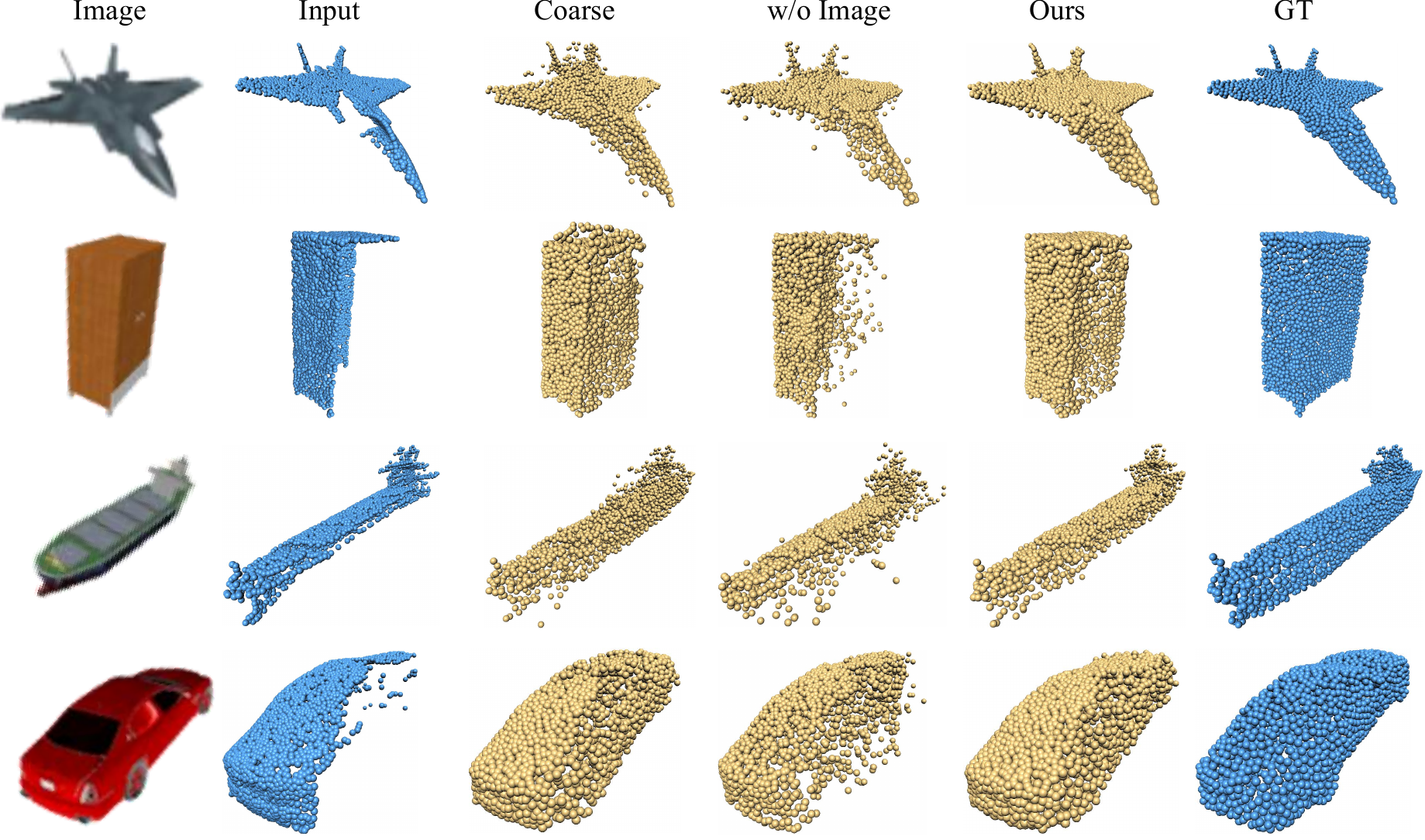}
  \caption{Visual comparisons between CSDN and its variants.}
  \label{fig:visualabla}
\end{figure*}

\subsubsection{Ablation on Shape Fusion}
To analyze the contribution of IPAdaIN, the comparison between three variants on the Shape Fusion module is shown in Table~\ref{tab:ablationSF}. We replace IPAdaIN in the variant A with a regular instance normalization. Without the global information provided by images, A cannot guarantee the completeness of the final shape and typically generates point clouds with higher CD and lower F-Score. The variant B, switching the positions of \emph{$F_P$} and \emph{$F_I$}, also produces points with higher CD and lower F-Score. In the variant C, we only utilize the image feature for the folding operation. Reconstructing shapes directly from images like \cite{zhang2021view} leads to more performance degradation. It further reveals the superiority of IPAdaIN over the single-view reconstruction method in terms of feature fusion. Moreover, from the similar performance of A and C, as well as the coarse result of CSDN, we can conclude that IPAdaIN plays the core role in the proposed cross-modal fusion strategy.

\subsubsection{Ablation on Dual-Refinement}
Table~\ref{tab:ablationDR} shows the comparison between the variants of CSDN on Dual-Refinement. The variant D, without the Local Refinement unit, leads to a large performance drop and coarser results. The variant E, without the Global Constraint unit, typically has lower performance. It means both the point cloud and image contribute to the refinement step, but the point cloud plays a major role. To understand the contribution of the double-refinement and dual-branch design, in the variant F we replace the parallel structure of dual-refinement with a serial structure (i.e., $P_0$ is sent to Local Refinement for the first refinement, and the refined result is sent to Global Constraint for final results). 
In contrast to the Dual-Refinement module,  F only uses information from both modalities for separate refinement without exploiting complementary information from each modality. The performance drop of F demonstrates the superiority of the proposed two-branch architecture.
\begin{table}
\tiny
    \renewcommand\arraystretch{1.3}
        \centering
        \caption{Comparisons between CSDN and its variants on Shape Fusion.}
        \label{tab:ablationSF}
        \small
        \begin{tabular}{c|c|c|c|c}
        \hline
        \multirow{2}{*}{Methods}& \multicolumn{2}{c}{Known} & \multicolumn{2}{c}{Novel} \cr\cline{2-5} & CD $\times 10^{-3}$ & F-Score & CD $\times 10^{-3}$ & F-Score \cr
        \hline
        \hline
                  w/o IPAdaIN (A) & 3.658 & 0.648 & 4.528 & 0.602 \\
                  \hline
                  SW (B) & 2.647 & 0.681 & 3.765 & 0.625 \\
                  \hline
                  ImgFolding (C) & 4.174 & 0.631 & 5.142 & 0.584 \\
                  \hline
                  Coarse & 3.752 & 0.632 & 6.757 & 0.548 \\
                  \hline
                  Ours (CSDN) & \textbf{2.570} & \textbf{0.695} & \textbf{3.656} & \textbf{0.631} \\
                  \hline
        \hline
        \end{tabular}
\end{table}
\begin{table}
    \renewcommand\arraystretch{1.3}
        \centering
        \caption{Comparisons between CSDN and its variants on Dual-Refinement.}
        \label{tab:ablationDR}
        \small
        \begin{tabular}{c|c|c|c|c}
        \hline
        \multirow{2}{*}{Methods}& \multicolumn{2}{c}{Known} & \multicolumn{2}{c}{Novel} \cr\cline{2-5} & CD $\times 10^{-3}$ & F-Score & CD $\times 10^{-3}$ & F-Score \cr
        \hline
        \hline
                  w/o LG (D) & 3.428 & 0.648 & 6.284 & 0.561 \\
                  \hline
                  w/o GC (E) & 2.700 & 0.687 & 4.015 & 0.624 \\
                  \hline
                  serial (F) & 3.036 & 0.659 & 4.227 & 0.611 \\
                  \hline
                  Coarse & 3.752 & 0.632 & 6.757 & 0.548 \\
                  \hline
                  Ours (CSDN) & \textbf{2.570} & \textbf{0.695} & \textbf{3.656} & \textbf{0.631} \\
                  \hline
        \hline
        \end{tabular}
\end{table}
\subsubsection{Ablation of Input Modality}
This set of ablative experiments aims at exploring the role of point clouds and images in our method. First, we change our network into single-modal versions and analyze the contribution of different input modalities. Specifically, we ablate on the single-view images in the variant G, where both IPAdaIN and Global Constraint are removed. The variant H is a single-view reconstruction network based on CSDN. We replace the shape fusion module with a vanilla foldingnet that uses only image features to reconstruct a coarse point cloud and remove the Local Refinement unit in the dual-refinement module. Then, we analyze their performance from two aspects: (1) missing geometry completion, and (2) original structure preservation.

\textbf{Missing Geometry Completion}
Table \ref{tab:ablationM} shows that the variants G and H have varying degrees of performance loss. While G still has competitive results on both known and novel categories, H has a large-margin performance drop (156\% on known categories and 231\% on novel categories).
Besides, these results further demonstrate that our method is an image-guided point cloud completion network, where point clouds play the major role.

We demonstrate some cases when images cannot depict the missing parts in point clouds and visually compare the completion results of CSDN and its variants G and H in Figure~\ref{fig:vis112}. More results under such a condition can be found in Figure~\ref{fig:morevis}. 
We can see that G struggles to infer the missing parts. Meanwhile, the single-view reconstruction H can only reconstruct coarse shapes with a considerable number of noisy points. Both the variants cannot recover the structures that are unseen in images (e.g., the wheels of the car, the upper part of the chair legs and the table legs). 
In contrast, CSDN can generate smoother detailed structures that are missing in both the input point clouds and images, rather than just reconstructing geometric details from images. We credit such superiority to our feature fusion strategy.

In Figure~\ref{fig:visualabla}, we show one more visual comparison between the final and coarse results of CSDN and its variant G. It can be observed that without the input images our method struggles to infer the missing regions and the coarse point clouds generated by Shape Fusion are already reasonably complete compared to the corresponding partial inputs.

\textbf{Original Structure Preservation}
To evaluate how our network preserves the initial structure of the objects, we also compare partial matching~\cite{wen2021cycle4completion} value between $P_{in}$ and the generated result $P_{out}$, which is single-side Chamfer Distance defined as
\begin{equation}
    \mathcal{PM}(P_{in},P_{out})=\frac{1}{\lvert P_{in} \rvert}\sum_{x \in P_{in}}\min_{y \in P_{out}} \lvert \lvert x-y \rvert \rvert.
\end{equation}
Table~\ref{tab:ablationPM} shows that the variant H struggles to preserve the partial shape since it uses only input images. 
However, our cross-modal network still has a better shape-preserving ability than the variant G that inputs only the partial point cloud. The comparison demonstrates that our method fuses cross-modal features to assist completion without losing the original shape information contained in point clouds. In other words, our method can efficiently extract and fuse the complementary information from the two modalities, rather than just reconstructing shapes from images. 
It is noteworthy that CSDN preserves initial points better than ViPC~\cite{zhang2021view}, which could also be concluded from Figures~\ref{fig:visualcomp} and~\ref{fig:morevis}.

\begin{figure}[h]
  \centering
  \includegraphics[width=\linewidth]{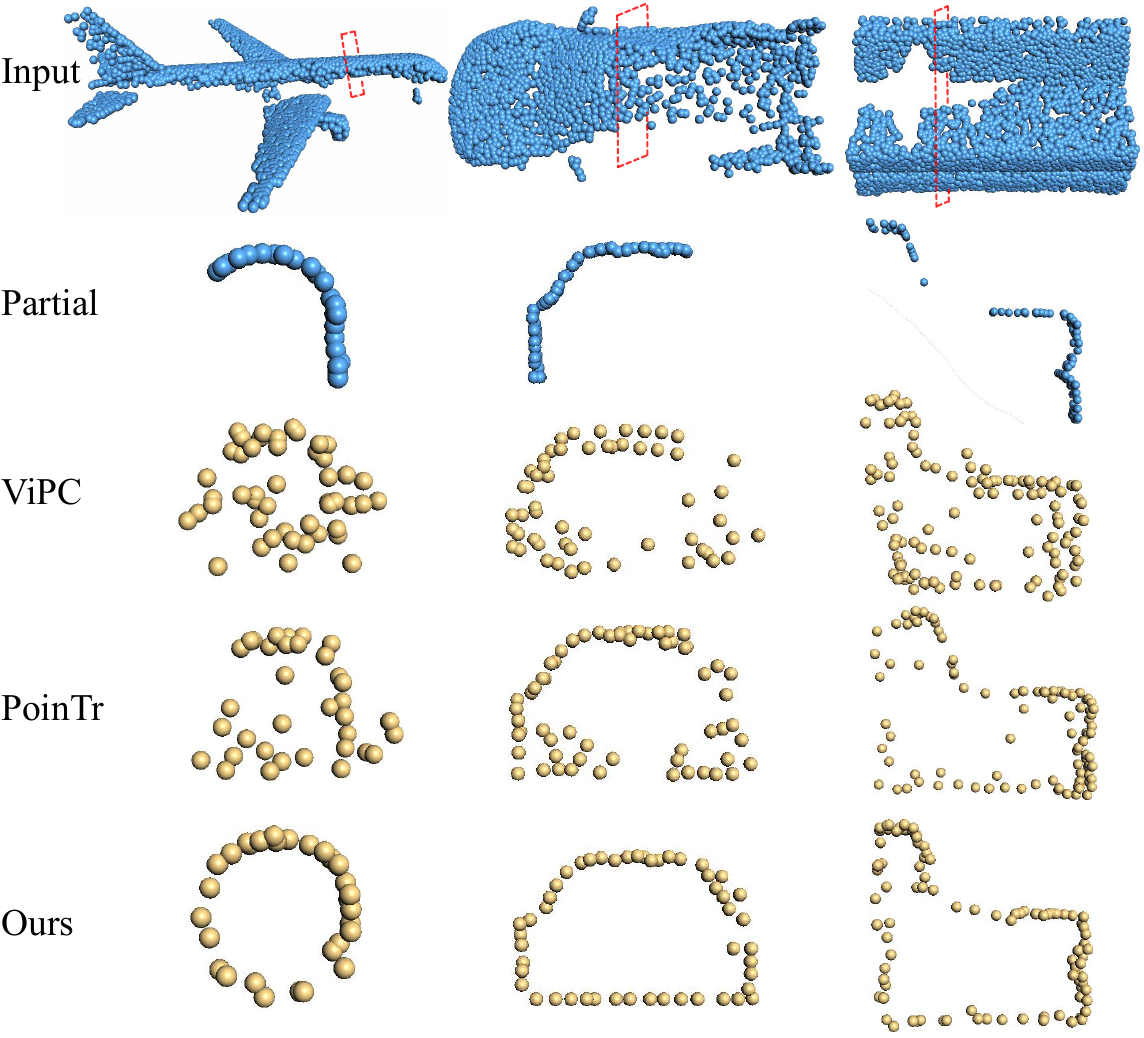}
  \caption{Visual comparisons of CSDN with \cite{zhang2021view,yu2021pointr} on structure preservation. Cross-sections of the completed point clouds are visualized. For each model, the 3D red rectangle in the first row indicates the position and orientation of the cross-section.}
  \label{fig:cut}
\end{figure}

It is also worth noting that CSDN can preserve slender structures (e.g., the rear wing of the car) better than the two variants using a single modality input in Figure~\ref{fig:vis112}. To better reveal its capability in structure preservation, thin cross-sections of some models are demonstrated in Figure~\ref{fig:cut}. We can see that CSDN preserves the original surfaces well while the other two competitors~\cite{zhang2021view,yu2021pointr} generate more noisy points inside the object.

\begin{table}
\tiny
    \renewcommand\arraystretch{1.3}
        \centering
        \caption{Comparisons between CSDN and its variants on the input modality.}
        \label{tab:ablationM}
        \small
        \begin{tabular}{c|c|c|c|c}
        \hline
        \multirow{2}{*}{Methods}& \multicolumn{2}{c}{Known} & \multicolumn{2}{c}{Novel} \cr\cline{2-5} & CD $\times 10^{-3}$ & F-Score & CD $\times 10^{-3}$ & F-Score \cr
        \hline
        \hline
                  w/o Image (G) & 4.179 & 0.629 & 4.974 & 0.588 \\
                  \hline
                  w/o Point (H) & 6.583 & 0.411 & 12.103 & 0.268 \\
                  \hline
                  Coarse & 3.752 & 0.632 & 6.757 & 0.548 \\
                  \hline
                  Ours (CSDN) & \textbf{2.570} & \textbf{0.695} & \textbf{3.656} & \textbf{0.631} \\
                  \hline
        \hline
        \end{tabular}
\end{table}

\begin{table}
\tiny
    \renewcommand\arraystretch{1.2}
        \centering
        \caption{Comparisons of Partial Matching value. Lower value means a better preserving ability. G and H refer to the variants in Table~\ref{tab:ablationM}.}
        \label{tab:ablationPM}
        \small
        \begin{tabular}{c|c|c|c|c|c}
        \hline
        Methods & Ours & G & H  & ViPC~\cite{zhang2021view} & PointAttN \cite{wang2022pointattn}\\
        \hline
        PM $\times 10^{-3}$ & 0.277 & 0.358 & 3.038 & 0.854 & 0.424 \\
        \hline
        \hline
        \end{tabular}
\end{table}
\begin{figure*}[h] 
  \centering
  \includegraphics[width=\textwidth]{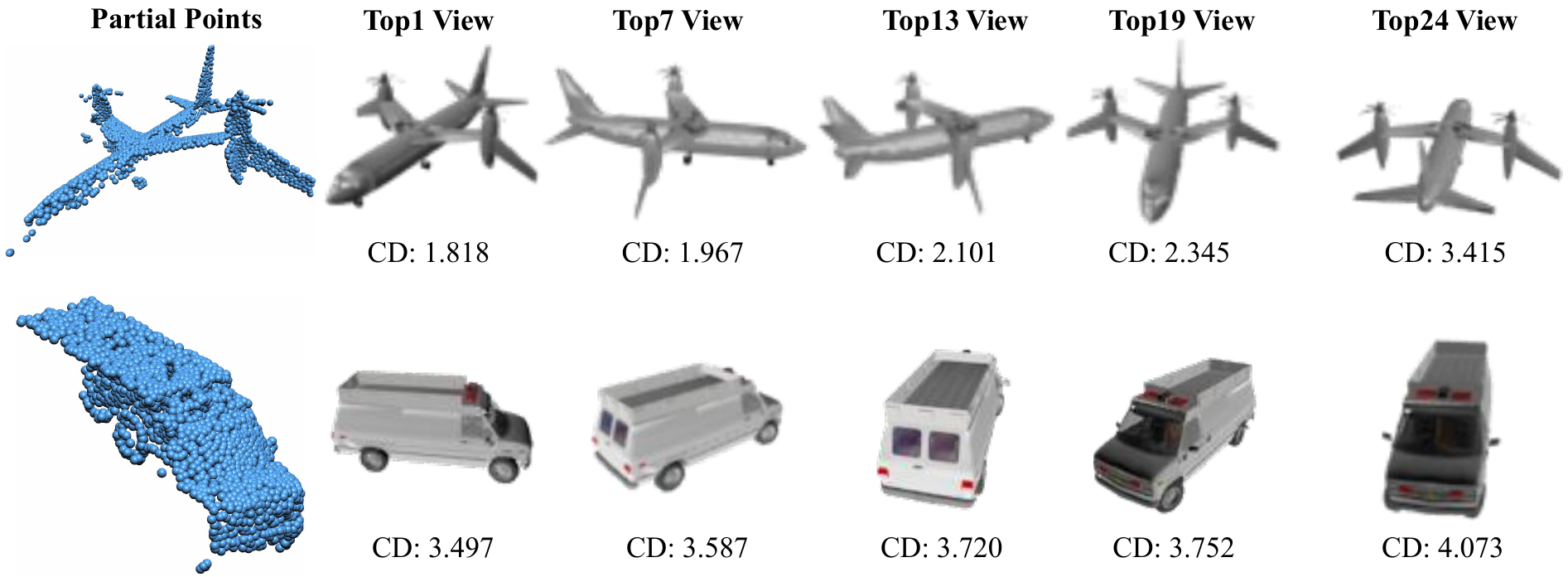}
  \caption{Images captured from different views often lead to slightly different completion results. A ``good" view that contains more complementary information for the input partial point clouds tends to produce the completion results with a lower CD value ($\times10^{-3}$).}
  \label{fig:viewEXP}
\end{figure*}
\begin{figure*}[ht]
  \centering
  \includegraphics[width=\textwidth]{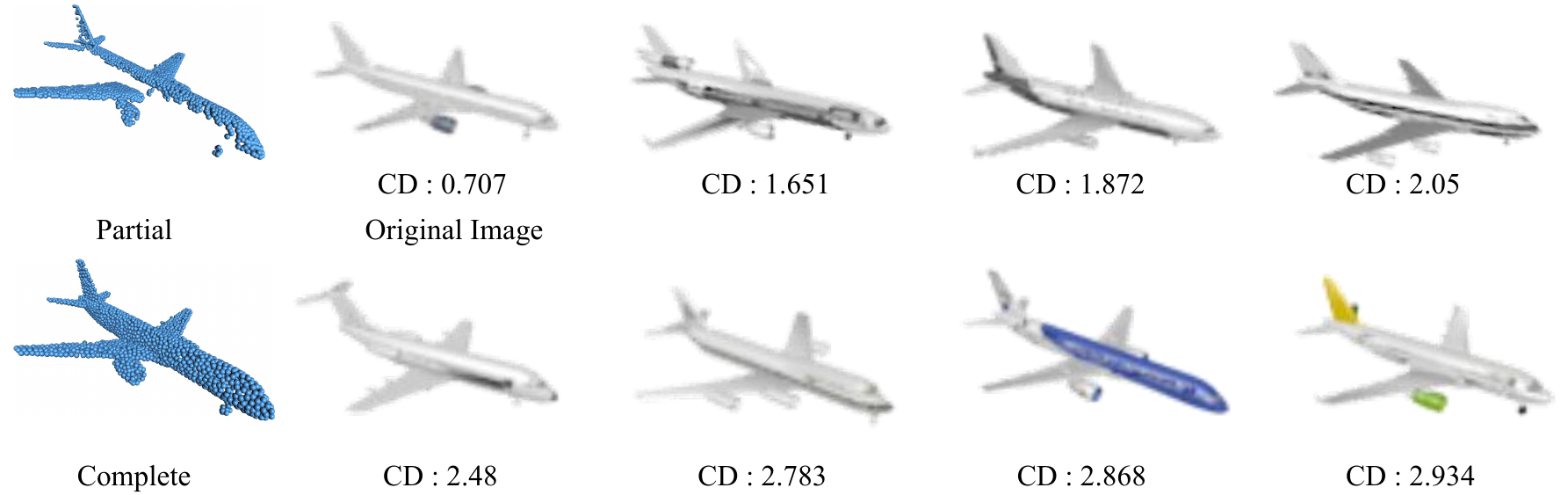}
  \caption{Results of using similar but intrinsically different images to bind the partial point cloud for completion. The object's structures and colors affect the completion results. The partial point cloud with its original image can infer a completion result with the smallest error of CD ($\times10^{-3}$).}
  \label{fig:similar}
\end{figure*}
\begin{figure*}[!t] 
  \centering
  \includegraphics[width=\textwidth]{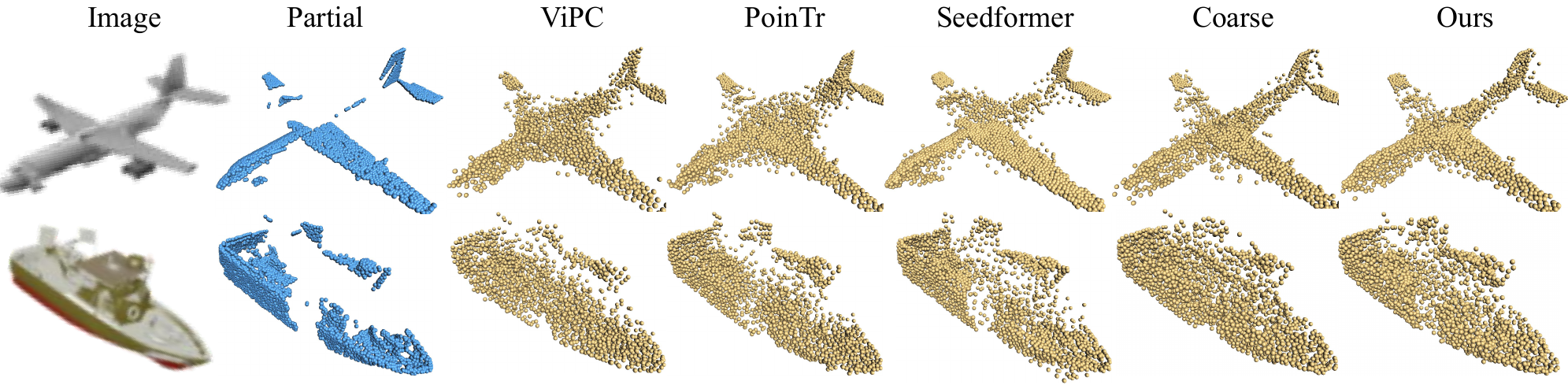}
  \caption{Failure cases. Similar to its competitors~\cite{zhang2021view,yu2021pointr,zhou2022seedformer}, our CSDN may generate poor completion results (the small structures cannot be reconstructed) when the partial point cloud lacks the main body and the assembled image has very low resolution. Please note that in such challenging cases, our CSDN can still recover the overall shape, which its competitors are not.}
  \label{fig:failure}
\end{figure*}
\begin{figure}[h]
  \centering
  \includegraphics[width=\linewidth]{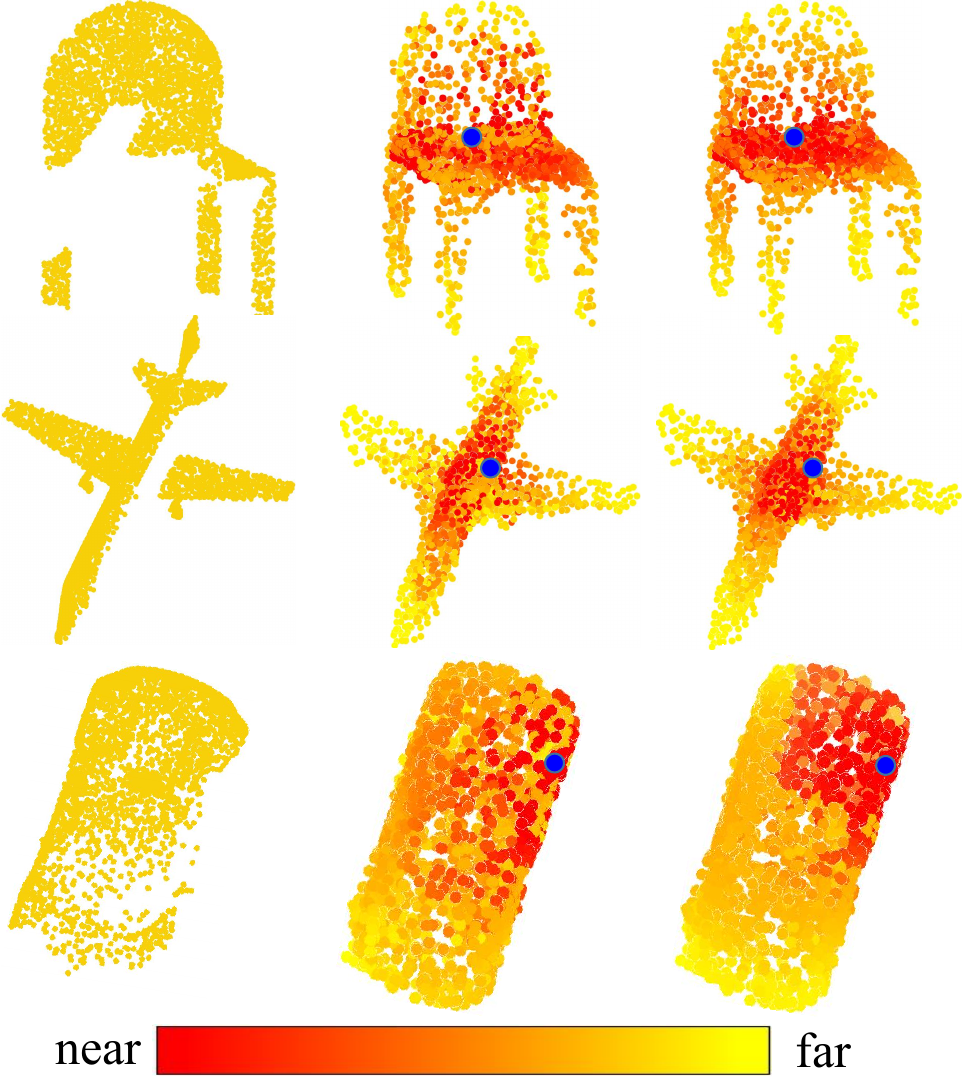}
  \caption{Visualization of Euclidean distances between the target points (in blue) and other points in the feature space. For each set, Left: Input partial point cloud; Middle: Distance before IPAdaIN; Right: Distance after IPAdaIN. The target points are randomly selected in missing parts.}
  \label{fig:feaeuvis}
\end{figure} 
\begin{figure}[h]
  \centering
  \includegraphics[width=\linewidth]{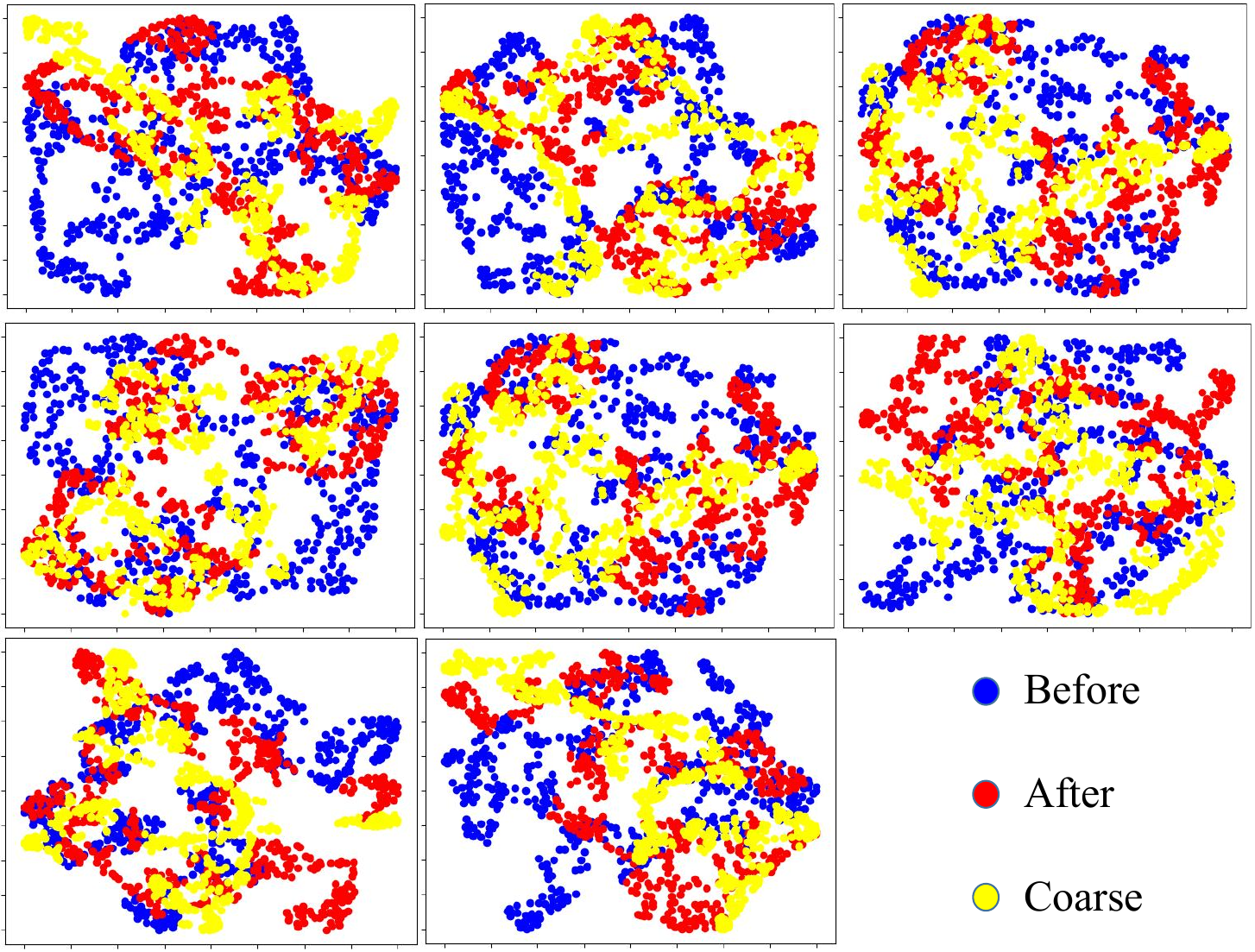}
  \caption{Visualization of feature distributions by t-SNE: $F_{before}$ (in blue), $F_{after}$ (in red) and $P_{coarse}$ (in yellow).}
  \label{fig:feavis}
\end{figure} 

\subsubsection{Ablation on Numbers of KNN}
We also conduct an ablation experiment on the number of nearest neighbors in the local refinement unit. $k$ is set to be 4, 8, 16, and 24, respectively. It is observed from Table \ref{tab:tab1} that CSDN has the best performance when $k=16$. 
However, we believe that this parameter needs to be fine-tuned for different datasets.
\begin{table}
\tiny
    \renewcommand\arraystretch{1.2}
        \centering
        \caption{Mean Chamfer distance with different point cloud resolutions. The first row denotes the number of input points.}
        \label{tab:resolution}
        \normalsize
        \begin{tabular}{c|c|c|c}
        \hline
        \multirow{2}{*}{Methods}& \multicolumn{3}{c}{Number of Input Points} \cr\cline{2-4} & 2048 & 1024 & 256 \cr
        \hline
        \hline
        Ours & 2.570 & 2.597 & 3.358  \\
        \hline
        PoinTr~\cite{yu2021pointr} & 2.851 & 2.882 & 4.255 \\
        \hline
        ViPC~\cite{zhang2021view} & 3.308 & 3.341 & 3.687 \\
        \hline
        \hline
        \end{tabular}
\end{table}
\begin{table}[ht]
    \tiny
    \renewcommand\arraystretch{1.2}
    \centering
    \caption{Quantitative comparisons with different KNN numbers.}
    \normalsize
    \begin{tabular}{c|c|c|c|c}
    \hline
    \multirow{2}{*}{Methods}& \multicolumn{2}{c}{Known} & \multicolumn{2}{c}{Novel} \cr\cline{2-5} & CD $\times 10^{-3}$ & F-Score & CD $\times 10^{-3}$ & F-Score \cr
    \hline
    \hline
              K = 4 & 2.643 & 0.691 & 4.061 & 0.625 \\
              \hline
              K = 8 & 2.636 & 0.688 & 4.053 & 0.622 \\
              \hline
              K = 16 & \textbf{2.570} & \textbf{0.695} & \textbf{3.656} & \textbf{0.631} \\
              \hline
              K = 24 & 2.656 & 0.685 & 4.116 & 0.624 \\
              \hline
    \hline
    \end{tabular}
    \label{tab:knn}
\end{table}
\begin{table*}
\tiny
    \renewcommand\arraystretch{1.2}
        \centering
        \caption{Standard deviations of CD on different-view images. The lower the value is,  the more view-independent the method is.}
        \label{tab:tab7}
        \normalsize
        \begin{tabular}{c|c|c|c|c|c|c|c|c|c}
        \hline
        {Methods}& 
        \multicolumn{9}{c}{Standard deviations of CD calculated on different images} \cr\cline{2-10}
        & Avg & Airplane & Cabinet & Car & Chair & Lamp & Sofa & Table & Watercraft \cr
        \hline
        \hline
                  ViPC \cite{zhang2021view} & 0.445 & 0.185 & 0.406 & 0.236 & 0.595 & 0.921 & 0.373 & 0.599 & 0.243 \\
                  \hline
                  Ours & \textbf{0.254} & \textbf{0.119} & \textbf{0.272} & \textbf{0.164} & \textbf{0.349} & \textbf{0.277} & \textbf{0.305} & \textbf{0.316} & \textbf{0.184} \\
                  \hline
        \hline
        \end{tabular}
\end{table*}
\begin{table}
    \renewcommand\arraystretch{1.3}
        \centering
        \caption{Results on different experiment settings. ``Different'' means images are randomly selected from 24 viewpoints. ``Same'' means images and partial point clouds are produced from the same viewpoints.}
        \label{tab:viewEXP}
        \small
        \begin{tabular}{c|c|c|c|c}
        \hline
        \multirow{2}{*}{Methods}& \multicolumn{2}{c}{Known} & \multicolumn{2}{c}{Novel} \cr\cline{2-5} & CD $\times 10^{-3}$ & F-Score & CD $\times 10^{-3}$ & F-Score \cr
        \hline
        \hline
                  Different & 2.570 & 0.695 & 3.656 & 0.631 \\
                  \hline
                  Same & 2.637 & 0.689 & 3.778 & 0.624 \\
                  \hline
        \hline
        \end{tabular}
\end{table}
\begin{table}
\tiny
    \renewcommand\arraystretch{1.2}
        \centering
        \caption{Comparisons on model sizes and performance.}
        \label{tab:size}
        \normalsize
        \begin{tabular}{c|c|c}
        \hline
        {Methods}&  Model size (Mib) & Performance (CD) \cr
        \hline
        \hline
                  ViPC \cite{zhang2021view} & 53.79 & 3.308 \\
                  \hline
                  PoinTr \cite{yu2021pointr} & 170.18 & 2.851 \\
                  \hline
                  Ours &  70.89 & 2.570 \\
                  \hline
        \hline
        \end{tabular}
\end{table}
\begin{table*}[h]
\centering
\caption{Summary of the limitations of our CSDN and its competitors.}
\renewcommand\arraystretch{1.2}
\small
\begin{tabularx}{16cm}{c|c}
\hline
Methods  & Limitation     \\ \hline \hline
     AltasNet\cite{groueix2018papier}
     & Rely on repeated reconstruction to represent a 3D shape as the union of several surface elements.
     \\ \hline 
     FoldingNet\cite{yang2018foldingnet}     
     & Overlook the local geometric characteristics.      \\ \hline
     PCN\cite{yuan2018pcn}
     & Only rely on a single global feature to generate points, hence incapable of synthesizing local details.                              \\ \hline 
     TopNet \cite{tchapmi2019topnet}                                         &  Results tend to have noisy points around the generated surface.
     \\ \hline
     PF-Net\cite{huang2020pf}                                                & Fail to recover local details to  certain extent.                   \\ \hline
     MSN\cite{liu2020morphing}
     & Lack of sufficient refinement and fail to generate fine-grained details.         
     \\ \hline  
     GRNet\cite{xie2020grnet}
     & \begin{tabular}[c]{@{}c@{}}It is subject to the resolution of the voxel representation,\\ which often leads to distortion in local complex structures.\end{tabular}
     \\ \hline
     PoinTr\cite{yu2021pointr}
     & The model has relatively more parameters owing to the transformer architecture.    
     \\ \hline
     ViPC\cite{zhang2021view}
     & \begin{tabular}[c]{@{}c@{}}Rely on the 3D reconstruction from a single image as well as an accurate 3D alignment.\\ The surface of the results is thus noisy.\end{tabular}
     \\ \hline
     Seedformer\cite{zhou2022seedformer}
     & The results are not evenly distributed.
     \\ \hline
     Ours 
     & Rely on the dual-domain training dataset. 
     \\ \hline \hline
\end{tabularx}%
\label{tab:limitations}
\end{table*}

\subsection{Impact of Point Cloud Resolution}
This section studies the stability of CSDN and two other competitors concerning point cloud resolution. 
To this end, we downsample the input point cloud to 1024 and 256 points respectively, and test the performance of the methods. From the results reported in Table~\ref{tab:resolution}, we observe that the performance drop of all three methods is small when the input is downsampled to 1024 points. This is because the point cloud can still convey the geometry of the partial shape. When the input is downsampled to 265 points, the performance of all methods degrades significantly. It is worth noting that the decline of ViPC (11.5\%) is smaller than that of CSDN (30.7\%) and PoinTr (49.2\%), which is due to that ViPC reconstructs points directly from images, while the input points play the major role in our method.

\subsection{Effect of Different-view Images}
\label{viewEXP}
This section studies the stability of CSDN when images can be collected from different views. Ideally, any good method of cross-modal point cloud completion should be view-independent. However, it is observed that images captured from different views may slightly affect the completion results. A ``good” view is considered to possess complementary information for the input partial point clouds. The images with a good view tend to produce the completion results with a lower CD value.

To validate the aforementioned statement, we randomly select some partial objects and test them with the individual images from all 24 views as input. As observed in Figure \ref{fig:viewEXP}, CSDN performs better on images that contain more information about the missing parts. 
For example, the image containing only the front or the end of a truck generates the results with the highest CD losses. 

To further demonstrate the stability of CSDN over the other cross-modal method, i.e. ViPC~\cite{zhang2021view}, we select 100 objects from each known category to evaluate how CSDN and ViPC perform with the images captured from different viewpoints. 
Specifically, we obtain 24 results for each incomplete object with the help of each image of the 24 different-view images.  We calculate the standard deviations for the 24 CD values to compare the stability of model performance, as shown in Table \ref{tab:tab7}.
The quantitative results clearly show that our CSDN  
is more view-independent than ViPC~\cite{zhang2021view} with a 42.9\% drop in the standard deviation.

In addition, we also test our method when images and the corresponding partial point clouds are acquired under the same viewpoints, with the results shown in Table~\ref{tab:viewEXP}. The negligible performance drop (2.6\% on the CD metric) indicates that CSDN is capable of extracting meaningful geometric information from images even when the images may not depict the missing parts of the objects. More visual comparisons can be found in Figure~\ref{fig:morevis}.

\subsection{Effect of Different Similar-images}
There may exist objects that have similar structures. Hence, we randomly select some models and replace the corresponding input images with similar images in the same category. 
Figure \ref{fig:similar} shows the results of one example. When inputting the pair of the original image and the partial point cloud, CSDN generates the completion result with a CD loss of $0.707 \times 10^{-3}$. 
Then, we replace the original image with seven images of the other models in ShapeNet-ViPC \cite{zhang2021view}. It is found that our method still works with similar images as input but produces results with a higher CD loss. Figure \ref{fig:similar} shows that the model performs better when the input image has similar structures and colors.
However, the last two images have nearly the same shape as the original image, but our method generates results with higher CD losses. We believe that it is caused by the remarkable differences in colors. Please note that we do not use such unpaired data during the training stage.
\subsection{Comparisons about Model Sizes}
We compare and report the model size of ViPC, PoinTr, and our CSDN in Table \ref{tab:size}. 
The comparison indicates that our CSDN has a slightly larger model size than ViPC but a much better performance, while PoinTr has the largest model size owing to its transformer architecture.

%% file: Feature_Visualization.tex
\subsection{Feature Visualization}
\begin{figure*}[h] 
  \centering
  \includegraphics[width=0.9\textwidth]{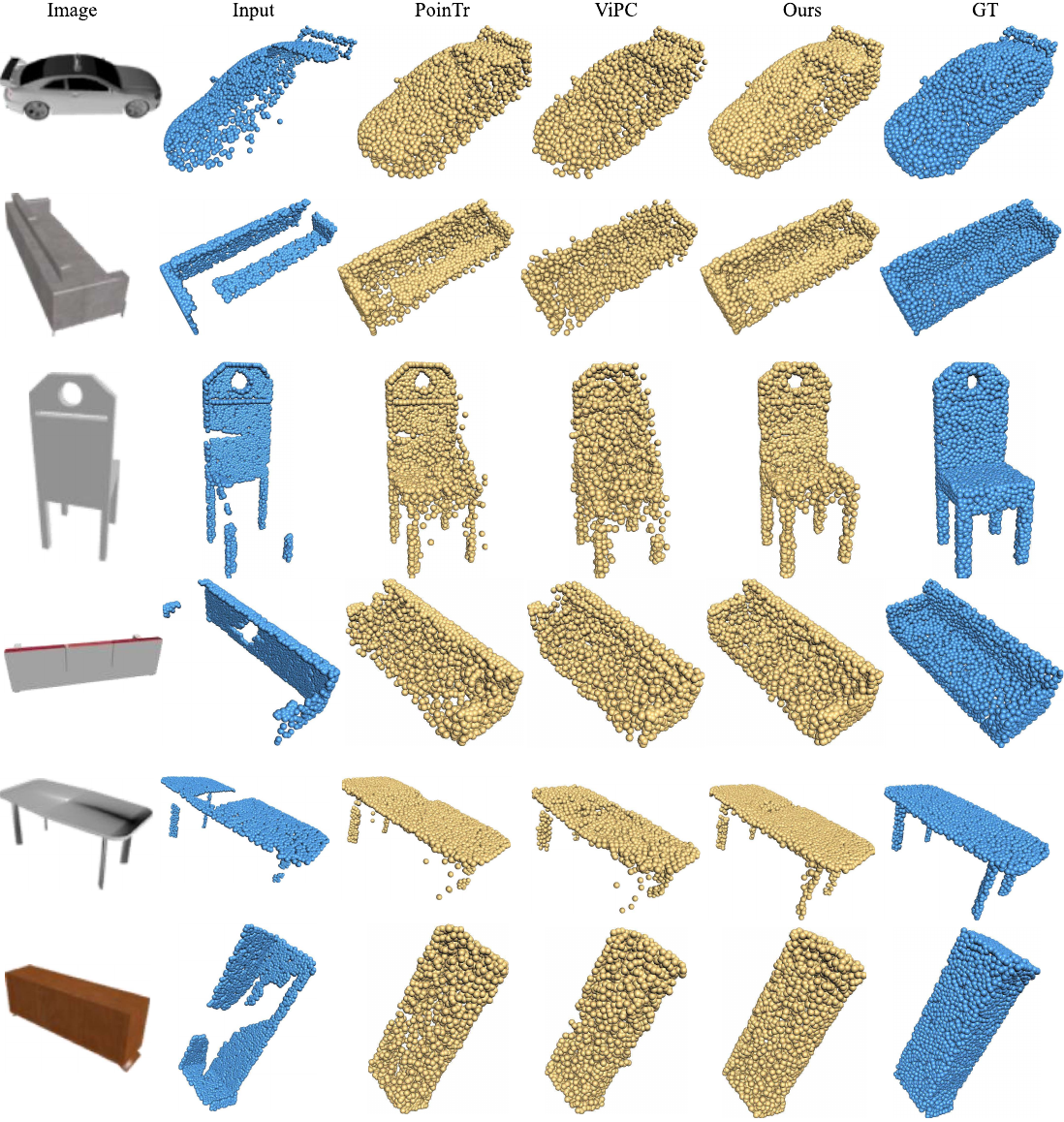}
  \caption{Comparisons between CSDN and its competitors~\cite{yu2021pointr,zhang2021view} when images and partial point clouds are captured under the same viewpoints.}
  \label{fig:morevis}
\end{figure*}

To better understand the role of IPAdaIN, this section provides visualizations of the features before and after IPAdaIN in the Shape Fusion module in two different ways. Specifically, we define the point-wise features before the last layer of IPAdaIN as $F_{before}=\{f_1,...,f_N\}\subseteq\mathbb{R}^C$, the corresponding point-wise features after IPAdaIN as $F_{after}$, and the generated coarse points as $P_{coarse}=\{p_1,...,p_N\}\subseteq\mathbb{R}^3$. Note that the affine transformation in IPAdaIN is the only step where image features are involved for point generation in the Shape Fusion module.

We first explore how features change before and after IPAdaIN. During the folding operation, points that are close in the 3D space should also be close in the feature space (i.e., $f_a$ should have a relatively small Euclidean distance to $f_b$ if $p_a$ and $p_b$ are spatially close). Figure~\ref{fig:feaeuvis} visualizes the coarse shapes $P_{coarse}$ generated by the Shape Fusion module, and the target points were randomly selected in the originally missing parts. Then, we compute the Euclidean distances between the target point and other points in the feature space. The visualization is generated by coloring points based on the distances. As shown in Figure~\ref{fig:feaeuvis}, points similar to the target point in the feature space are spread across the entire point cloud before IPadaIN. After IPadaIN, the similar points in the feature space become closer in the 3D space. It should be noted that IPAdaIN does not involve any learnable parameters other than the affine parameters generated by the image features. This means that the affine transformation of IPAdaIN changes the feature distributions and normalizes the generated points to the complete shape.

In addition, we plot $F_{before}$, $F_{after}$, and $P_{coarse}$ as 2D images by t-SNE~\cite{van2008visualizing} in Figure \ref{fig:feavis} to disclose the change in feature distribution after IPAdaIN. It can be concluded that IPAdaIN makes the distribution of point-wise feature $F_{after}$ more similar to its direct result $P_{coarse}$ than $F_{before}$. These results further explain that IPAdaIN fuses modalities by changing feature statistics.

%% file: limitation.tex
\section{Limitations and Failure Cases}
Table~\ref{tab:limitations} summarizes the limitations among these compared methods to provoke further research insights. Besides, we visualize some failure cases whose CD losses are rather larger than the average error of that category, as shown in Figure \ref{fig:failure}. Although the shape fusion step succeeds in recovering the complete shape to a certain extent, and the dual-refinement module also removes noisy points, the final results are still less pleasing. This is especially the case when the partial point clouds have large missing regions both in their main bodies and detailed structures. Meanwhile, the input images are of very low resolutions, which can only provide limited complementary information for the recovery of fine geometries (e.g., the ship cabin). This problem could be solved by training on a dataset that has real-world high-resolution images and dense point clouds. Besides, although the motivation of the proposed CSDN is to reconstruct better shapes with the help of single-view images, multi-view images and neural rendering techniques would also help point cloud completion, which is our future work.

%% file: conclusion.tex
\section{Conclusion}
We present a novel point cloud completion paradigm that mimics the physical object-repairing process. The paradigm is implemented as a cross-modal shape-transfer double-refinement network (CSDN). Experiments on standard point cloud completion benchmarks demonstrate the significant improvement of our CSDN over the pioneer multi-modal method ViPC and the other SOTAs.
Our work has revealed that the fusion of features from cross-modalities facilitates the generation of reliable geometries to complete partial point clouds. This benefits from the fact that features learned from images not only constrain the generation of the overall shape of an object but also help to refine its local details, indicating the potential of using images to recover finer geometry of objects from partial point clouds.
